\documentclass[conference]{IEEEtran}
\IEEEoverridecommandlockouts

\let\cal\mathcal

\newcommand{\rbr}[1]{\left(#1\right)}
\newcommand{\mlpsmall}{\text{\scriptsize{MLP}}}
\newcommand{\linsmall}{\text{\scriptsize{LIN}}}

\usepackage{stfloats}

\usepackage{booktabs}
\usepackage{cite}
\usepackage{amsmath,amssymb,amsfonts}

\usepackage{algorithm,algpseudocode}
\usepackage{placeins}

\usepackage{subcaption}

\usepackage{multirow}
\usepackage{graphicx}
\usepackage{textcomp}
\usepackage{comment}
\usepackage{xcolor}
\usepackage{bm}
\def\BibTeX{{\rm B\kern-.05em{\sc i\kern-.025em b}\kern-.08em
    T\kern-.1667em\lower.7ex\hbox{E}\kern-.125emX}}
\begin{document}

\title{ClusterViG: Efficient Globally Aware \\Vision GNNs via Image Partitioning}

\newcommand\blfootnote[1]{%
  \begingroup
  \renewcommand\thefootnote{}\footnote{#1}%
  \addtocounter{footnote}{-1}%
  \endgroup
}

\author{
\IEEEauthorblockN{Dhruv Parikh\IEEEauthorrefmark{1}, Jacob Fein-Ashley\IEEEauthorrefmark{1}, Tian Ye\IEEEauthorrefmark{1}, Rajgopal Kannan\IEEEauthorrefmark{2}, Viktor Prasanna\IEEEauthorrefmark{1}} 
\IEEEauthorblockA{
    \IEEEauthorrefmark{1}University of Southern California \IEEEauthorrefmark{2}DEVCOM Army Research Office \\
    \IEEEauthorrefmark{1}\{dhruvash, feinashl, tye69227, prasanna\}@usc.edu \IEEEauthorrefmark{2}rajgopal.kannan.civ@army.mil
}
}

\maketitle

\begin{abstract}
Convolutional Neural Networks (CNN) and Vision Transformers (ViT) have dominated the field of Computer Vision (CV). Graph Neural Networks (GNN) have performed remarkably well across diverse domains because they can represent complex relationships via unstructured graphs. However, the applicability of GNNs for visual tasks was unexplored till the introduction of Vision GNNs (ViG). Despite the success of ViGs, their performance is severely bottlenecked due to the expensive $k$-Nearest Neighbors ($k$-NN) based graph construction. Recent works addressing this bottleneck impose constraints on the flexibility of GNNs to build unstructured graphs, undermining their core advantage while introducing additional inefficiencies. To address these issues, in this paper, we propose a novel method called Dynamic Efficient Graph Convolution (DEGC) for designing efficient and globally aware ViGs. DEGC partitions the input image and constructs graphs in parallel for each partition, improving graph construction efficiency. Further, DEGC integrates local intra-graph and global inter-graph feature learning, enabling enhanced global context awareness. Using DEGC as a building block, we propose a novel CNN-GNN architecture, ClusterViG, for CV tasks. Extensive experiments indicate that ClusterViG reduces end-to-end inference latency for vision tasks by up to $5\times$ when compared against a suite of models such as ViG, ViHGNN, PVG, and GreedyViG, with a similar model parameter count. Additionally, ClusterViG reaches state-of-the-art performance on image classification, object detection, and instance segmentation tasks, demonstrating the effectiveness of the proposed globally aware learning strategy. Finally, input partitioning performed by DEGC enables ClusterViG to be trained efficiently on higher-resolution images, underscoring the scalability of our approach. ClusterViG enables real-time deployment of GNN-based models for CV, providing a new avenue for future research.  
\end{abstract}


\section{Introduction}
\label{sec: intro}

\blfootnote{\textbf{Distribution Statement A:} Approved for public release. Distribution is unlimited.}

Graph Neural Networks (GNN) have demonstrated superior performance across diverse domains ranging from complex network analysis \cite{qiu2018deepinf, zhang2018link}, knowledge graphs \cite{zhou-etal-2019-gear, qiu-etal-2019-dynamically}. Recommender systems \cite{wang2019kgat, wang2020multi} to chemistry \cite{cho2018three}, physical systems \cite{hoshen2017vain, pmlr-v80-sanchez-gonzalez18a} and brain network analysis \cite{ktena2017distance}. GNNs utilize unstructured graphs comprising nodes (vertices) and edges (links) to represent data. This enables their performance on domains involving non-Euclidean data, using graphs to represent such complex data to perform feature representation learning at the node/edge/graph-level features \cite{zhou2022graph}. Despite their success in the above domains, the applicability of GNNs across Computer Vision (CV) has been limited.

Traditionally, CV has been dominated by Convolutional Neural Network (CNN) based architectures \cite{alexnet, efficientnet}. CNNs have been widely used for image classification \cite{resnet}, object detection \cite{Cai_2018_CVPR}, semantic segmentation \cite{fcn_semantic}, instance segmentation \cite{instance_seg}, etc. tasks in CV. Recently, Multi-Layer Perceptron (MLP) based vision models \cite{tolstikhin2021mlp, touvron2022resmlp} have also performed well across several CV tasks. CNN and MLP-based models treat the input image as a structured grid of pixels and apply structured convolution and fully connected layers to the pixel grid for feature representation learning \cite{alexnet, tolstikhin2021mlp}.

The transformer architecture, initially introduced for Natural Language Processing (NLP) in \cite{vaswani2017attention}, has significantly influenced advancements in CV \cite{dosovitskiy2021an, li2022uniformer, liu2022swin, carion2020end, chen2021topological}. Vision Transformers (ViT) \cite{dosovitskiy2021an} represent an input image as a sequence of patches and utilize the transformer architecture to process such a sequence. Unlike CNNs and MLPs, which operate with local receptive fields, Vision Transformers (ViTs) leverage the Multi-Headed Self-Attention (MHSA) mechanism to achieve a global receptive field \cite{greedyvig}. 

While CNN, MLP, and ViT-based models have been successful in CV, they are limited by their reliance on rigid grid or sequence representations of images. Vision GNNs (ViG), proposed in \cite{vig}, address this limitation by using graphs to represent images, offering greater flexibility. Additionally, ViG utilizes GNNs to learn about such image graphs, extending the GNN framework to CV. 

Despite its superior performance across benchmark CV tasks, ViG suffers from two severe drawbacks: (i) High computational expense of dynamic graph construction. Specifically, ViG employs $k$-Nearest Neighbors ($k$-NN) to connect similar image patches (nodes) with edges, forming an image graph. This $k$-NN graph construction is repeated at every model layer as patch (node) features are updated through the GNN layers. (ii) Limited flexibility of image graph construction. Specifically, ViG restricts edge construction in the image graph to pairwise, first-order similar nodes (patches), which reduces the model's receptive field and undermines the primary advantage of using unstructured graphs to represent images.

Several recent works \cite{vihgnn, greedyvig, pvg, mobilevig, scalemobilevig} have been proposed to address the above two drawbacks of ViG. ViHGNN \cite{vihgnn} and PVG \cite{pvg} increase the flexibility of image graph construction while increasing its computational expense. Meanwhile, GreedyViG \cite{greedyvig} and MobileViG \cite{mobilevig, scalemobilevig} utilize structured graphs for image graph construction, reducing computational expense at the cost of flexibility. Thus, no prior work exists that addresses \emph{both} drawbacks of ViG \cite{vig} and ViG-based models \cite{vihgnn, greedyvig, pvg, mobilevig, scalemobilevig}, \emph{simultaneously}.

In light of this, in this paper, we propose a novel method called Dynamic Efficient Graph Convolution (DEGC) to design a new class of ViG-based models, ClusterViG. ClusterViG significantly reduces the computational expense of dynamic image graph construction \emph{without} sacrificing its flexibility. DEGC, the main building block of ClusterViG, facilitates this improvement. DEGC partitions the input image patches and performs graph construction in parallel for each partition, reducing the computational expense of dynamic image graph construction. It then learns a global feature vector for each partition through inter-graph interactions among the constructed graphs. Finally, each partition's local patch (node) features are updated in parallel, using \emph{both} local and global features through intra-graph interactions. This enhances the receptive field of ClusterViG and fully leverages the flexibility of unstructured graph representation learning via GNNs. We summarize the contributions of our work below,

\begin{itemize}
    \item We propose a novel method, Dynamic Efficient Graph Convolution (DEGC), to build a new class of ViG models, ClusterViG. ClusterViG reduces the computational expense of dynamic image graph construction \emph{without} sacrificing its flexibility for graph representation learning via GNNs. 
    
    \item DEGC is a general method for designing efficient Graph Convolutional Networks (GCNs) for CV. We demonstrate its generality by implementing DEGCs efficient dynamic image graph construction and flexible local-global graph representation learning with standard GCN backbones.
    
    \item ClusterViG reaches state-of-the-art (SOTA) performance across three representative CV tasks: ImageNet image classification \cite{imagenet}, COCO object detection \cite{cocotasks}, and COCO instance segmentation \cite{cocotasks}, when compared to its SOTA counterparts (CNN, MLP, ViT or ViG based), with comparable total model parameters and GMACs.
    
    \item ClusterViG reaches this SOTA performance with a hardware-friendly \emph{isotropic} architecture. This contrasts with prior SOTA models, which typically have a pyramid architecture.

    \item ClusterViG achieves up to $5\times$ reduction in end-to-end inference latency and throughput for vision tasks, validated on a SOTA heterogeneous CPU-GPU platform.

    \item ClusterViG enables training ViG models on images with up to $2\times$ higher resolution ($4\times$ more nodes) by leveraging efficient dynamic graph construction through input image partitioning. In contrast to prior ViG models, which are hindered by high graph construction costs, ClusterViG achieves superior scalability.
\end{itemize}

The remainder of this paper is organized as follows: Sec. \ref{sec: related} reviews related work in computer vision, encompassing both ViG and non-ViG based architectures. Sec. \ref{sec: prelim} provides a formal introduction to GNNs and GCNs, ending with a clear problem definition. Sec. \ref{sec: method} elaborates on the proposed DEGC method, including its optimizations and the detailed ClusterViG architecture. Sec. \ref{sec: exp} presents experimental results, covering both benchmark computer vision tasks and the inference performance of ClusterViG, concluding with Sec. \ref{sec: concl}.

\section{Related Work}
\label{sec: related}

\subsection{CNN, MLP and ViT in CV} 
\label{subsec: related-cnn-mlp-vit-cv}
CNNs have dominated CV \cite{resnet, denseconnectedcnn, imagenet, fcn_semantic}. Despite the success of ViTs, their high computational cost (quadratic in total image patches/tokens) has prevented their deployment in real-time edge applications \cite{vit-survey, dosovitskiy2021an}. Subsequently, CNNs have thrived in the space of efficient CV \cite{mobilenet, mobilenetv2, efficientnet, efficientnetv2}. Recently, several ViT-based architectures have been proposed that aim to reduce the computational cost of ViTs \cite{fast-vit-hilo, A-vit, efficientvit, efficientformer}. Further, hybrid CNN-ViT architectures have also been proposed that effectively capture local and global information \cite{rethinking-vit, separable-vit-self-att, fast-vit-struct-reparam}. Based on ViT, MLP-based architectures have been proposed that are efficient and perform competitively on benchmark vision tasks \cite{touvron2022resmlp, tolstikhin2021mlp, cyclemlp, phase-vision-mlp}. While the above CNN, MLP, and ViT-based architectures have been successful in CV, they are restricted to representing an image as either a rigid grid or a sequence, limiting their learning capability \cite{vig}.

\subsection{GNN in CV}
\label{subsec: related-gnn-cv}
Graphs are flexible, general-purpose data structures. Graph representation learning, facilitated by GNN, has demonstrated SOTA performance across social, biological, citation, or brain networks (graphs) \cite{graphsage, gcn-kipf, ktena2017distance, gnn-survey-2020}. The application of GNNs to CV was restricted to applications such as point cloud classification and segmentation \cite{edgeconv, mrgcn, geomgcn, agcn}, scene graph generation \cite{scenegraphgeneration}, human action recognition \cite{gnn-human-action}, etc. \cite{gnn-in-vision-survey}. GNNs were not used as general-purpose backbones for vision tasks until the introduction of ViG \cite{vig}. Subsequently, several works have been proposed to address the drawbacks of ViG \cite{vihgnn, pvg, greedyvig, mobilevig, scalemobilevig}, advancing GNN-based backbone models for CV. Vision HyperGraph Neural Network (ViHGNN) \cite{vihgnn} represents an image as a hypergraph and uses HyperGraph Neural Networks (HGNN) for feature learning. Hypergraphs offer greater flexibility than graphs in representing inter-patch information, as their edges can connect multiple nodes (image patches), not just pairs. Despite its added flexibility, ViHGNN is more computationally expensive than ViG, as constructing an image hypergraph incurs higher costs than $k$-NN based image graph construction. Progressive Vision Graph (PVG) \cite{pvg} introduces Progressively Separated Graph Construction (PSGC) for second-order node similarity, along with a novel aggregation-update module (MaxE) and activation function (GraphLU) in its GCN layer. PVG outperforms ViG on CV benchmarks but shares the identical drawback of higher computational expense as ViHGNN. MobileViG \cite{mobilevig} and MobileViGv2 \cite{scalemobilevig} were proposed to address the computation expense of image graph construction in ViG-based models, enabling their deployment on mobile devices. However, the proposed graph construction strategy in MobileViG and MobileViGv2, Sparse Vision Graph Attention (SVGA), and Mobile Graph Convolution (MGC) are fully static. Static graph construction connects image patches in a structured fashion, irrespective of patch features, based on their spatial location within the image. Thus, while reducing the computational expense of graph construction, such methods severely restrict the flexibility of graph representation learning. GreedyViG \cite{greedyvig} proposes a dynamic version of SVGA, Dynamic Axial Graph Construction (DAGC), to address the limitations of MobileViG \cite{mobilevig, scalemobilevig}. The proposed DAGC starts with a structured graph constructed via SVGA. Then, it prunes connections in SVGA based on image patch similarity. GreedyViG has two drawbacks: (i) The DAGC graph is initialized from the static SVGA graph, inherently limiting graph construction flexibility. Any image patch connection missed by SVGA due to its static nature is also excluded from the constructed DAGC graph. (ii) DAGCs forward pass relies on a \emph{for} loop with an expensive \emph{roll} operation to extract image patch connections across the image, significantly increasing the computational expense of GreedyViG and degrading its inference performance.

\begin{figure}
    \centering
    \includegraphics[width=\linewidth]{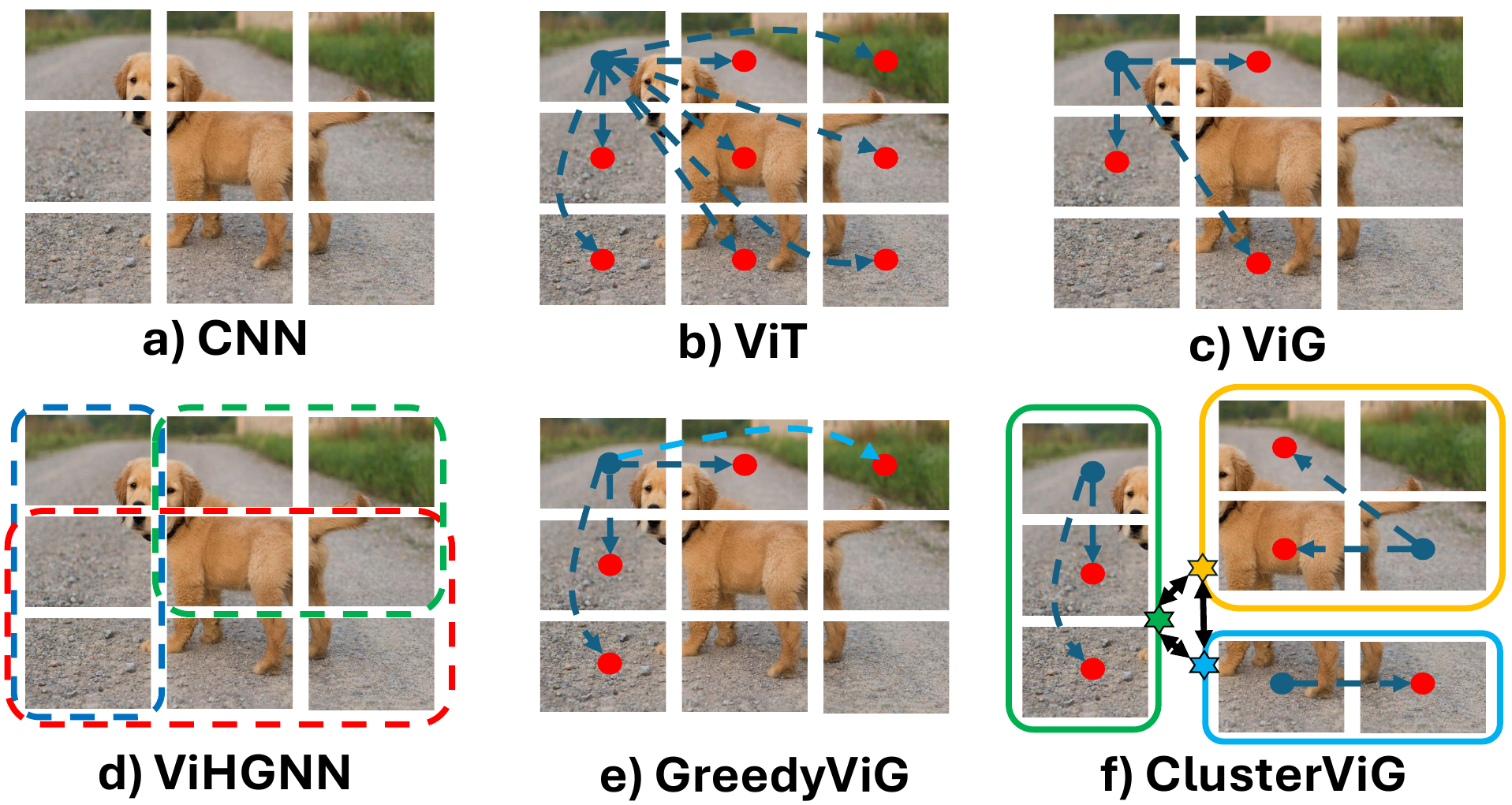}
    \caption{Comparing methods. a) CNN treats an image as a grid of patches; b) ViT treats an image as a sequence of patches and computes a dense attention matrix that compares each patch (blue) to all other patches (red); c) ViG connects a patch to similar patches, via $k$-NN; d) ViHGNN uses a hypergraph to represent image patches (a hyperedge contains several patches/nodes, shown as colored boxes); e) GreedyViG uses SVGA to connect a patch to its spatial horizontal and vertical axis patches and prunes connections (light blue); e) ClusterViG partitions the patches for fast graph construction and performs local-global feature learning (global features shown via star). 
    }
    \label{fig: method_comparison_summary}
\end{figure}

Compared to prior works, our work focuses on improving the inference performance of ViG-based models without sacrificing the flexibility of graph representation learning. DEGC-enabled ClusterViG is faster and more flexible than \cite{vig, vihgnn, pvg, greedyvig, mobilevig, scalemobilevig}, addressing the fundamental drawbacks concerning ViG-based models. Fig. \ref{fig: method_comparison_summary} visually compares representative approaches.

\section{Preliminaries}
\label{sec: prelim}
In this section, we start by formally defining a GNN \cite{mpgnn}. Next, we briefly describe the main GCN variants generally used in CV \cite{edgeconv, mrgcn, gin, graphsage}. Finally, we conclude with a formal definition of a problem for building general-purpose vision backbones using GNNs.

\subsection{GNN}

\textbf{Graph.} A graph is defined as $\cal{G}(\cal{V}, \cal{E})$ where $\cal{V}$ is the set of nodes (vertices) and $\cal{E}$ is the set of edges (links) in the graph $\cal{G}$. Specifically, $\cal{E} = \{(j,i) \;|\; j, i \in \cal{V} \; \text{and} \; j \rightarrow i \; \text{is a directed edge in }\cal{G}\}$. Further, each node $i \in \cal{V}$ has an associated (node) feature vector $\bm{x}_i \in \mathbb{R}^{D}$. The (node) feature matrix $\bm{X} \in \mathbb{R}^{N \times D}$ contains feature vectors of all the nodes, where the node set $\cal{V} = \{0, 1, \ldots, N-1\}$ and $\bm{x}_{i \in \cal{V}} = \bm{X}(i, :)$.

\textbf{Graph Neural Network (GNN).} Given an input graph $\cal{G}(\cal{V}, \cal{E})$ with a node feature matrix $\bm{X}$, an $L$ layered GNN can be described using the message passing framework as below \cite{mpgnn},

\begin{equation} \label{eq: mp-gnn}
    \bm{x}_i^{(l)} = \Psi^{(l)}\rbr{\bm{x}_i^{(l-1)}, \bigoplus_{j \in \cal{N}(i)} \Phi^{(l)}\rbr{\bm{x}_i^{(l-1)}, \bm{x}_j^{(l-1)}}} \\
\end{equation}

Note that Eq. \ref{eq: mp-gnn} is applied $\forall i \in \cal{V}$ and $\forall l \in \{1, 2, \ldots, L\}$. The input to the $l^{\text{th}}$ GNN layer is the output of the previous layer, $\bm{X}^{(l-1)} \in \mathbb{R}^{N \times D^{(l-1)}}$. For $l=1$, we have, $\bm{X}^{(0)} = \bm{X}$ and $D^{(0)} = D$. $\cal{N}(i) = \{j \;|\; (j, i) \in \cal{E}\}$ is the set of neighboring nodes for node $i$. Depending on the type of GNN, $\cal{N}(i)$ may contain $i$ (self-loop). $\bigoplus$ is an aggregation operation, such as sum, max, or mean. Lastly, $\Psi^{(l)}$ and $\Phi^{(l)}$ are learnable functions, such as MLPs. 

Eq. \ref{eq: mp-gnn} describes how the feature vector of every node $i \in \cal{V}$ is updated at a layer $l$ of the GNN, yielding the updated node feature matrix $\bm{X}^{(l)} \in \mathbb{R}^{N\times D^{(l)}}$. We can further decompose Eq. \ref{eq: mp-gnn} into three core operations: (i) \emph{Message Creation}. For each node $j$ connected to node $i$ via edge $(j,i)$, a message $\bm{m}_{j,i}^{(l)} = \Phi^{(l)}\rbr{\bm{x}_i^{(l-1)}, \bm{x}_j^{(l-1)}}$ is created. (ii) \emph{Aggregation}. The messages are aggregated over $\cal{N}(i)$ as $\bm{a}_i^{(l)} = \bigoplus_{j\in \cal{N}(i)} \bm{m}_{j,i}^{(l)}$. (iii) \emph{Update}. Finally, the feature vector of node $i$ is updated as $\bm{x}_i^{(l)} = \Psi^{(l)}\rbr{\bm{x}_i^{(l-1)}, \bm{a}_i^{(l)}}$.   

\begin{table}[]
    \centering
    \caption{Node feature vector update rule for GCN variants}
    \vspace{-5pt}
    \resizebox{\columnwidth}{!}{
    \begin{tabular}{cc}
         \toprule
         GCN &  Feature Vector Update Rule (Node $i$, Layer $l$)\\
         \toprule
         EdgeConv & $\bm{x}_i^{(l)} = \max_{j \in \cal{N}(i)} \mlpsmall^{(l)}\rbr{\bm{x}_i^{(l-1)} \;|\; \bm{x}_j^{(l-1)} - \bm{x}_i^{(l-1)}}$\\
         \midrule
         MRGCN & $ \bm{x}_i^{(l)} = \mlpsmall^{(l)} \left( \bm{x}_i^{(l-1)} \; \middle| \; \max_{j \in \cal{N}(i)} \rbr{\bm{x}_j^{(l-1)} - \bm{x}_i^{(l-1)}} \right)$ \\
         \midrule
         GraphSAGE & $\bm{x}_i^{(l)} =\mlpsmall_{\Psi}^{(l)} \left( \bm{x}_i^{(l-1)} \;\middle|\; \max_{j \in \cal{N}(i)} \mlpsmall_{\Phi}^{(l)}\rbr{\bm{x}_j^{(l-1)}} \right)$\\
         \midrule
         GIN & $\bm{x}_i^{(l)} = \mlpsmall^{(l)} \rbr{(1 + \epsilon) \bm{x}_i^{(l-1)} + \rbr{\sum_{j \in \cal{N}(i)} 
         \bm{x}_j^{(l-1)}}}$\\
         \bottomrule
    \end{tabular}
    }
    \label{tab: gcn_variant_update_rule}
\end{table}

\textbf{Graph Convolutional Network (GCN).} GCNs are a subset of GNNs commonly used in CV tasks \cite{vig, mrgcn}. Specifically, GCN architectures such as EdgeConv \cite{edgeconv}, MRGCN \cite{mrgcn}, GraphSAGE \cite{graphsage}, and GIN \cite{gin} serve as foundational components to build ViG-based backbone models for CV. In Table \ref{tab: gcn_variant_update_rule}, we summarize the update rule for the feature vector of a node $i$, at layer $l$, for the four GCN variants, utilizing the general message passing framework for GNNs. Note that $|$ refers to the concatenation operation, $\mlpsmall^{(l)}$ refers to the MLP used at layer $l$ of the GCN, and $\epsilon$, for GIN, is a learnable scalar parameter.
Additionally, note that $\cal{N}(i)$ includes node $i$ (self-loop) for all GCN variants in Table \ref{tab: gcn_variant_update_rule}.

\subsection{Problem Definition}
\label{subsec: prelim_problem_def}
We now formally define the problem, motivating our method. Let $\bm{I} \in \mathbb{R}^{H \times W \times C}$ represent an image patch feature grid, where $H$ and $W$ denote the grid dimensions, and $C$ is the length of the feature vector for each patch. The image patch feature grid, $\bm{I}$, is spatially flattened (tokenized) to $\bm{X} \in \mathbb{R}^{N \times C}$, where $N = HW$ represents the total number of image patches. The central problem in constructing a ViG-based backbone model now lies in defining a \emph{module} that can perform the following two operations: (i) \emph{Graph Construction.} The \emph{module} first constructs a graph $\cal{G}(\cal{V}, \cal{E})$ via a mapping $\cal{F}$, based on the image patch (node) features $\bm{X}$ as $\cal{F}: \bm{X} \mapsto \cal{G}(\cal{V}, \cal{E})$. (ii) \emph{Graph Representation Learning.} The \emph{module} now uses a single-layer\footnote{A single \emph{module} comprises a single-layer GCN. The \emph{module} is repeated in a structured manner to construct ViG-based backbone models.} GCN to update the patch (node) features via a mapping $\cal{H}_{\Theta}: \rbr{\cal{G}(\cal{V}, \cal{E}), \bm{X}} \mapsto \bm{X}'$. Note that $\bm{X}' \in \mathbb{R}^{N \times C'}$ represents the updated patch (node) features with $C'$ output feature length. Further note that $\cal{H}_\Theta$ represents the GCN, parameterized by $\Theta$. Finally, the updated patch features are reshaped to $\bm{I}' \in \mathbb{R}^{H \times W \times C'}$ for subsequent processing. 

This learnable \emph{module} serves as the building block to construct a general-purpose ViG-based backbone model, with several instances of the \emph{module} used to facilitate multi-layer, multi-hop graph representation learning via GNNs. In ViG, the mapping $\cal{F}$ is simply the $k$-NN operation, which constructs a graph by connecting each image patch to its $k$ most similar patches. ViG uses the GCN variants in Table \ref{tab: gcn_variant_update_rule} for $\cal{H}_\Theta$.
\section{Method}
\label{sec: method}

\begin{figure*}
    \centering
    \includegraphics[width=\linewidth]{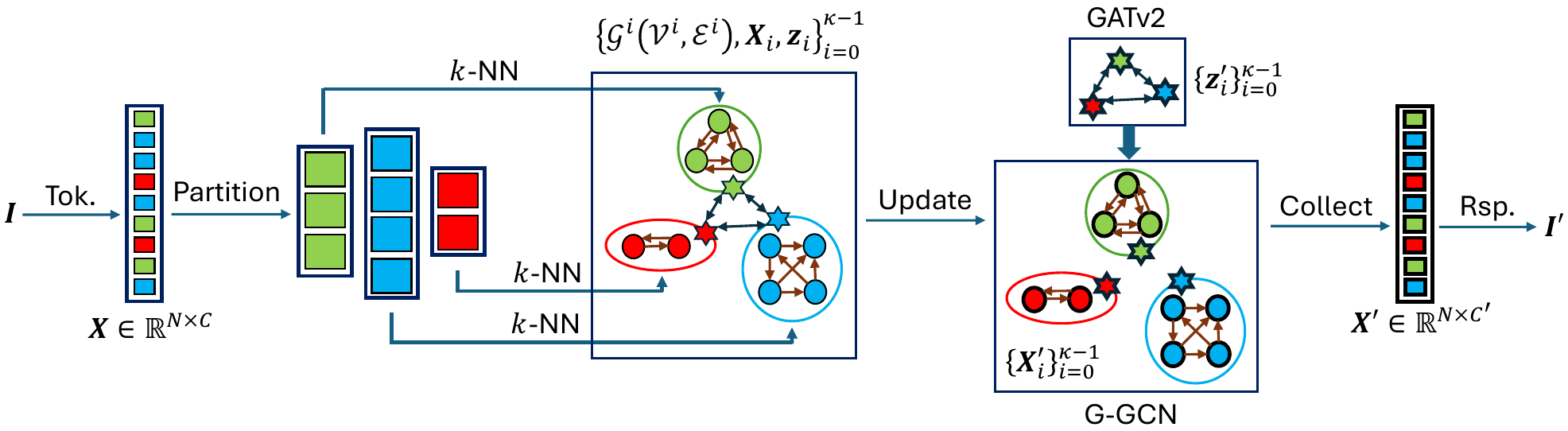}
    \caption{Dynamic Efficient Graph Convolution \emph{module}. DEGC performs four main operations: (i) Input Partitioning. $\bm{X}$ is partitioned into $\kappa$ partitions via clustering. (ii) Graph Construction. A $k$-NN graph is constructed in each partition, in parallel across all partitions. (iii) Global Update. Each partition's global feature vector, $\bm{z}_i$, is updated via GATv2 through global inter-partition interactions, yielding $\bm{z}_i'$. (iv) Local Update. Each partition updates its patch features by incorporating $\bm{z}_i'$, via intra-partition interactions, in parallel, yielding $\bm{X}'$, which is reshaped to $\bm{I}'$. Note that \textbf{dark borders} indicate updated features.}
    \label{fig: degc}
\end{figure*}

We propose a novel \emph{module}, Dynamic Efficient Graph Convolution (DEGC), to build a new class of ViG-based backbone models, ClusterViG. The proposed DEGC \emph{module} defines a computationally efficient mapping $\cal{F}$ by partitioning the image patches and constructing graphs in parallel for each partition. Next, DEGC defines $\cal{H}_\Theta$, incorporating local-global feature learning on the constructed graphs via GNNs. Notably, DEGC does not \emph{trade-off} the flexibility of graph representation learning to reduce the computational expense of image graph construction, unlike prior works.

In this section, we detail the DEGC \emph{module}, outline optimizations for fast inference, and describe the ClusterViG architecture constructed using DEGC as the building block.

\subsection{DEGC}
The operations in the DEGC \emph{module} can be decomposed into the following four steps: (i) Input Partitioning, (ii) Graph Construction, (iii) Global Update, and (iv) Local Update. We describe each step in detail.


\textbf{Input Partitioning.} The input image patch feature grid, $\bm{I} \in \mathbb{R}^{H \times W \times C}$, is first tokenized to $\bm{X} \in \mathbb{R}^{N \times C}$, where $N = HW$. The $N$ tokens (patches) are treated as nodes, $\cal{V} = \{0, 1, \ldots, N-1\}$, with node (patch) feature matrix $\bm{X}$. We now partition $\cal{V}$ and $\bm{X}$ into $\kappa$ partitions, where the $i^{\text{th}}$ partition has a node set $\cal{V}^i$ with a corresponding node feature matrix $\bm{X}_i \in \mathbb{R}^{|\cal{V}^i| \times C}$, for $0 \leq i \leq \kappa - 1$. This partitioning is achieved using \( k \)-Means clustering, where the number of partitions, \( \kappa \), corresponds to the total number of clusters. Let node $v$ belong to a cluster (partition) $i \in \{0, 1, \ldots, \kappa - 1\}$, i.e., $v \in \cal{V}^i$. Since a cluster contains nodes with similar features, we assume that the \( K \) nearest neighbors of the node \( v \), \( \mathcal{N}_K(v) \), lie within \( \mathcal{V}^i \), i.e., \( \mathcal{N}_K(v) \subseteq \mathcal{V}^i \). Consequently, this assumption reduces the search space for \( k \)-NN graph construction for any node \( v \in \mathcal{V}^i \subset \mathcal{V} \) to its partition \( \mathcal{V}^i \), significantly lowering computational expense while maintaining flexibility. This naturally motivates our use of clustering for partitioning. 



\textbf{Graph Construction.} Based on the assumption $\cal{N}_K(v) \subseteq \cal{V}^i$, we perform $k$-NN graph construction in each partition locally, connecting a node $v \in \cal{V}^i$ to its $K$ nearest neighbors (based on node feature similarity) in $\cal{V}^i$. Since graph construction is performed locally in each partition, it can be easily and efficiently parallelized. Consequently, this reduces the cost of $k$-NN graph construction from $\cal{O}\rbr{N^2}$ to $\cal{O}\rbr{\frac{N^2}{\kappa^2}}$, assuming an equal number of nodes per partition. After graph construction, we obtain $\kappa$ graphs with corresponding node feature matrices, $\{ (\cal{G}^i(\cal{V}^i, \cal{E}^i), \bm{X}_i)\}_{i=0}^{\kappa - 1}$.


\textbf{Global Update.} We first compute a global feature vector for each partition to enable global feature learning. Specifically, for a partition $i$, we compute its global feature vector as, $\bm{z}_i = \frac{1}{|\cal{V}^i|} \mathbf{1}^{\top}_{|\cal{V}^i|} \bm{X}_i$, for all $i \in \{0, 1, \ldots, \kappa - 1\}$. Here, $\bm{1}_{|\cal{V}^i|} \in \mathbb{R}^{|\cal{V}^i|}$ is a vector of $1$'s of length $|\cal{V}^i|$. Note that $\bm{z}_i$ is the centroid for the $i^{\text{th}}$ partition. The $\kappa$ centroids of all the partitions, $\{ \bm{z}_i \}_{i=0}^{\kappa-1}$, are used to construct a global graph, $\cal{G}^{\cal{C}}(\cal{V}^\cal{C}, \cal{E}^\cal{C})$. This global graph $\cal{G}^{\cal{C}}$ has node set $\cal{V}^{\cal{C}} = \{0, 1, \ldots, \kappa-1\}$ and edge set $\cal{E}^{\cal{C}} = \{(j,i) \;|\; j, i \in \cal{V}^{\cal{C}}\}$. Further note that $\cal{G}^{\cal{C}}$ is a directed, complete graph with self-loops, where each partition centroid is connected to all other centroids, including itself. The node (global) feature matrix for $\cal{G}^{\cal{C}}$ is $\bm{Z} \in \mathbb{R}^{\kappa \times C}$, where $\bm{z}_i = \bm{Z}(i, :)$. We use a single-layer Graph Attention Network, GATv2 \cite{gatv2}, to perform global feature learning on $\cal{G}^{\cal{C}}$ via the global feature matrix $\bm{Z}$. Specifically, the global feature vector of the $i^{\text{th}}$ partition, $\bm{z}_i$, is updated as follows,

\begin{equation}
\label{eq: z_update}
\bm{z}_i' = \sum_{j=0}^{\kappa - 1} \alpha_{ji} \linsmall_{\text{\tiny{L}}} (\bm{z}_j)
\end{equation}
where,
\begin{equation}
\label{eq: att_score}
\alpha_{ji} = \frac{\exp{\rbr{\bm{a}^\top \text{LeakyReLU} \rbr{\linsmall_{\text{\tiny{R}}}(\bm{z}_{i}) +  \linsmall_{\text{\tiny{L}}}(\bm{z}_j) } }}} {\sum_{k=0}^{\kappa-1}\exp{\rbr{\bm{a}^\top \text{LeakyReLU} \rbr{ \linsmall_{\text{\tiny{R}}}(\bm{z}_{i}) +  \linsmall_{\text{\tiny{L}}}(\bm{z}_k) } }}}
\end{equation}

In Eq. \ref{eq: z_update} and \ref{eq: att_score}, $\bm{z}_i' \in \mathbb{R}^{\Tilde{C}}$ is the updated global feature vector. $\linsmall_{\text{\tiny{L}}}(.)$ and $\linsmall_{\text{\tiny{R}}}(.)$ are learnable linear projection layers that project a vector in $\mathbb{R}^{C}$ to $\mathbb{R}^{\Tilde{C}}$. Finally, $\bm{a} \in \mathbb{R}^{\Tilde{C}}$ is a learnable vector. The global feature vector for partition $i$ is updated as a weighted sum of the linearly projected global feature vectors of all partitions, including itself (Eq. \ref{eq: z_update}) where the weights (attention scores) are computed via Eq. \ref{eq: att_score}. Thus, this framework aggregates information from all partitions, weighted by their learned importance, into each partition's global feature vector. Consequently, DEGC captures global inter-partition relationships, enhancing its ability to represent and utilize contextual dependencies.


\textbf{Local Update.} We now have $\{(\cal{G}^i(\cal{V}^i, \cal{E}^i), \bm{X}_i, \bm{z}_i')\}_{i=0}^{\kappa-1}$, after the above steps, where each partition $i$ has a globally learned feature vector, $\bm{z}_i'$. We propose Global-GCN (G-GCN), a novel GCN that incorporates each partition's globally learned feature vector to enrich the local features of its patches (nodes). Specifically, the update rule for a single-layer G-GCN is given as follows,

\begin{equation} \label{eq: g-gcn-single-layer}
    \begin{split}
    \bm{x}_v' = \Psi\rbr{\bm{x}_v, \bm{z}_i', \bigoplus_{u \in \cal{N}(v)} \Phi\rbr{\bm{x}_v, \bm{x}_u, \bm{z}_i'}} 
    \end{split}
\end{equation}

Note that Eq. \ref{eq: g-gcn-single-layer} is used to update the local features for all the patches (nodes) within a partition, i.e., $\forall v \in \cal{V}^i$, across all partitions, i.e., $\forall i \in \{0, 1, \ldots, \kappa-1\}$. Further, note that the local update step can be easily and efficiently parallelized across all the partitions since each partition works on its patches (nodes). We leverage the globally-aware local update of G-GCN (Eq. \ref{eq: g-gcn-single-layer}) and extend it to four GCN variants, resulting in G-EdgeConv, G-MRGCN, G-GraphSAGE, and G-GIN. Table \ref{tab: g-gcn-variant-update-rule} gives the single-layer update equation for these G-GCN variants. Note that $\delta$ (for G-GIN) is a scalar learnable parameter. Incorporating $\bm{z}_i'$ to update each patch (node) in partition $i$ enriches its feature vector with global information, enhancing DEGCs overall learning capacity. The updated patch features from each partition yield the updated feature matrix $\bm{X}' \in \mathbb{R}^{N \times C'}$, which is then reshaped into $\bm{I}' \in \mathbb{R}^{H \times W \times C'}$ for subsequent processing.

Finally, we note that each DEGC \emph{module} executes the four steps described above. Notably, a DEGC \emph{module} dynamically computes partitions and constructs graphs based on the current node feature matrix every time it is applied. This repeated dynamic computation enables adapting graph representations flexibly for efficient local-global feature learning.

\begin{table}[]
    \centering
    \caption{Node feature vector update rule for G-GCN variants (single layer)}
    \vspace{-5pt}
    \resizebox{\columnwidth}{!}{
    \begin{tabular}{cc}
        \toprule
         G-GCN & Feature Vector Update Rule (Node $v$, Partition $i$) \\
        \toprule
         G-EdgeConv & $\bm{x}_v' = \max_{u \in \cal{N}(v)} \mlpsmall \rbr{\bm{x}_v \;|\; \bm{x}_u -\bm{x}_v \;|\; \bm{z}_i'}$ \\
         \midrule
         G-MRGCN & $\bm{x}_v' = \mlpsmall \left( \bm{x}_v \;\middle|\; \max_{u \in \cal{N}(v)}\rbr{\bm{x}_u - \bm{x}_v} \;\middle|\; \bm{z}_i' \right) $ \\
         \midrule
         G-GraphSAGE & $\bm{x}_v' = \mlpsmall_\Psi \left( \bm{x}_v \;\middle|\; \max_{u \in \cal{N}(v)} \mlpsmall_\Phi (\bm{x}_u) \;\middle|\;  \bm{z}_i' \right) $ \\
         \midrule 
         G-GIN & $\bm{x}_v' = \mlpsmall \rbr{(1+\epsilon)\bm{x}_v  + \rbr{\sum_{u \in \cal{N}(v)} \bm{x}_u} + (1+\delta)\bm{z}_i'}$ \\
         \bottomrule
    \end{tabular}}
    \label{tab: g-gcn-variant-update-rule}
\end{table}

\begin{algorithm}
    \caption{Dynamic Efficient Graph Convolution (DEGC)}
    \label{alg: degc}
    \begin{algorithmic}[1]
        
        \Require $\bm{I} \in \mathbb{R}^{H \times W \times C}$, image patch feature grid; $\kappa$, total partitions; $K$, neighbors per partition.  
        \Ensure $\bm{I}' \in \mathbb{R}^{H \times W \times C'}$, updated image patch feature grid.

        \State{$\bm{X} \gets \text{tokenize}(\bm{I})$} \Comment{$\bm{X} \in \mathbb{R}^{N \times C}, N = HW$}

        \State{$\cal{V} \gets \{0, 1, \ldots, N-1\}$}

        \State{$\{ (\cal{V}^{i}, \bm{X}_i) \}_{i=0}^{\kappa - 1} \gets \text{partition}(\cal{V}, \bm{X})$}    \Comment{$\kappa$ partitions}

        \For{$i = 0, \ldots, \kappa - 1 \;\textbf{parallel}$}
            \State{$\cal{G}^{i}(\cal{V}^{i}, \cal{E}^{i}) \gets k\text{-NN}(\cal{V}^{i}, \bm{X}_i, K)$}  \Comment{parallel graph construction}
        \EndFor

        \For{$i = 0, \ldots, \kappa - 1$}
            \State{$\bm{z}_i \gets \frac{1}{|\cal{V}^{i}|} \mathbf{1}_{|\cal{V}^{i}|}^{\top} \bm{X}_i$} \Comment{$\bm{z}_i \in \mathbb{R}^C$, global feature}
        \EndFor

        \For{$i = 0, \ldots, \kappa -1$}
            
            \State{$\bm{z}_i'\gets \sum_{j=0}^{\kappa - 1} \alpha_{ji} \linsmall_{\text{\tiny{L}}} (\bm{z}_j)$} \Comment{global feature update}

            \State{where, $\alpha_{ji} = \frac{\exp{\rbr{\bm{a}^\top \text{LeakyReLU} \rbr{ \linsmall_{\text{\tiny{R}}}(\bm{z}_{i}) 
            +  \linsmall_{\text{\tiny{L}}}(\bm{z}_j) } }}} 
            {\sum_{k=0}^{\kappa-1}\exp{\rbr{\bm{a}^\top \text{LeakyReLU} \rbr{ \linsmall_{\text{\tiny{R}}}(\bm{z}_{i}) 
            +  \linsmall_{\text{\tiny{L}}}(\bm{z}_k) } }}} 
            $}
            
        \EndFor

        \For{$i = 0, \ldots, \kappa - 1$ \textbf{parallel}}
            
            \State{// $i^{\text{th}}$ partition $(\cal{G}^{i}(\cal{V}^{i}, \cal{E}^{i}), \bm{X}_i, \bm{z}_i')$}

            \For{$v \in \cal{V}^{i}$}
                \State{$\bm{x}_v' = \Psi\rbr{\bm{x}_v, \bm{z}_i', \bigoplus_{u \in \cal{N}(v)} \Phi\rbr{\bm{x}_v, \bm{x}_u, \bm{z}_i'}  }  $ } 
            \EndFor
            
        \EndFor

        \State{$\bm{X}' \gets \{\bm{x}_v' \;|\; v \in \cal{V}\}$\footnotemark} \Comment{$\bm{X}' \in \mathbb{R}^{N \times C'}$}

        \State{$\bm{I}' \gets \text{reshape}(\bm{X}')$} \Comment{$\bm{I}' \in \mathbb{R}^{H \times W \times C'}$}
        \\
        \Return{$\bm{I}'$}
        
    \end{algorithmic}   
\end{algorithm}
\footnotetext{We slightly abuse notation here as $\bm{X}'$ is a tensor, not a set. $\bm{X}'$ contains the updated patch feature vectors for all patches across all partitions.}

\subsection{Optimizations}

The DEGC \emph{module} achieves two key improvements: (i) it reduces the $k$-NN graph construction cost by a factor of $\kappa^2$ by parallelizing graph construction across partitions, and (ii) it enables parallel local patch feature updates within partitions. However, two key challenges arise in implementing DEGC optimally: (i) the overhead of partitioning and (ii) the creation of uneven partitions. We propose an optimized DEGC \emph{module} implementation that effectively addresses both challenges.

\begin{figure}
    \centering
    \includegraphics[width=\linewidth]{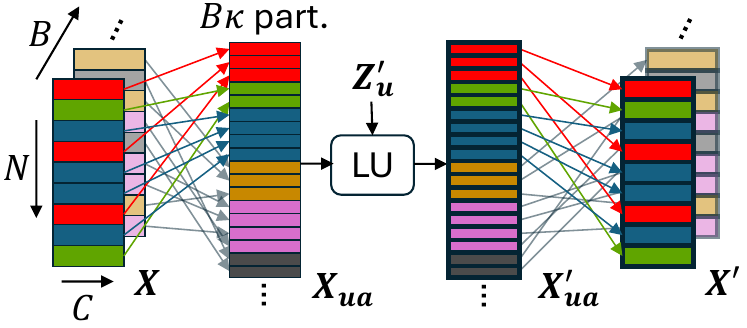}
    \caption{Optimized DEGC \emph{module}. LU refers to $\text{local\_update()}$ function, and the \textbf{dark borders} indicate updated patch features.}
    \label{fig: scatter_gather_batch}
\end{figure}

\begin{algorithm}
    \caption{Optimized DEGC \emph{module}}
    \label{alg: optimal_degc}
    \begin{algorithmic}[1]
        \Require{$\bm{X} \in \mathbb{R}^{B \times N \times C}$}
        \Ensure{$\bm{X}' \in \mathbb{R}^{B \times N \times C'}$}

        \State{$\bm{Z}, \bm{\mathrm{labels}} \gets \text{batched\_cluster}(\bm{X})$} \Comment{$B \times \kappa \times C, B \times N$}

        \State{$ \bm{Z}_u' \gets \text{global\_update}(\bm{Z}) $} \Comment{$B\kappa \times \Tilde{C}$}

        \State{$\bm{X}_{ua}, \bm{\mathrm{batch}} \gets \text{scatter}(\bm{X}, \bm{\mathrm{labels}})$} \Comment{$BN\times C, BN$}

        \State{$\bm{X}_{ua}' \gets \text{local\_update}(\bm{X}_{ua}, \bm{Z}_u', \bm{\mathrm{batch}})$} \Comment{$BN \times C'$ \hspace{1em}($B\kappa$ partitions with $BN$ total nodes processed in parallel)}
        
        \State{$\bm{X}' \gets \text{gather}(\bm{X}_{ua}', \bm{\mathrm{labels}})$} \Comment{$B \times N \times C'$}
        \\

        \Return{$\bm{X}'$}

    \end{algorithmic}
\end{algorithm}


\textbf{Partitioning Overhead.} The partitioning overhead arises from clustering \( N \) patches via \( k \)-Means. The computational complexity of \( k \)-Means is linear with respect to \( N \) (for a small number of total clusters, \( \kappa \)), in contrast to \( k \)-NN, which scales quadratically. To ensure computational efficiency, we restrict \( k \)-Means to a single random initialization and limit the total number of iterations to 20. For ClusterViG variants, we constrain \( \kappa \) to a maximum of 6. Additionally, to support efficient batched inference and training, we implement a batched \( k \)-Means clustering algorithm capable of processing multiple images simultaneously within a batch.

\textbf{Uneven Partitions.} $k$-Means clustering typically produces balanced clusters \cite{zhou2020effect_kmeans}, but the resulting partitions are often uneven. Algorithm \ref{alg: optimal_degc} addresses handling uneven partitions. Given a batched input $\bm{X} \in \mathbb{R}^{B \times N \times C}$, centroids $\bm{Z} \in \mathbb{R}^{B \times \kappa \times C}$ and cluster labels $\bm{\mathrm{labels}} \in \mathbb{R}^{B \times N}$ are computed using $\text{batched\_cluster}()$. Here, $\bm{Z}$ holds the batch-wise centroids, and $\bm{\mathrm{labels}}$ assigns a cluster label in $\{0, 1, \ldots, \kappa-1\}$ to each patch in $\bm{X}$. The $\text{global\_update}()$ function updates the centroids to $\bm{Z}_u' \in \mathbb{R}^{B\kappa \times \Tilde{C}}$, with the batch dimension flattened (\emph{unrolled}\footnote{Unrolling flattens the batch dimension: $\bm{A} \in \mathbb{R}^{B \times N \times C}$ becomes $\bm{A}_u \in \mathbb{R}^{BN \times C}$.}). The input $\bm{X}$ is scattered into an \emph{unrolled} and \emph{arranged} tensor $\bm{X}_{ua} \in \mathbb{R}^{BN \times C}$ using $\text{scatter}()$ via $\bm{\mathrm{labels}}$. As shown in Fig. \ref{fig: scatter_gather_batch}, $\bm{X}_{ua}$ holds image partitions arranged along the unrolled batch dimension. The tensor $\bm{\mathrm{batch}} \in \mathbb{R}^{BN}$ assigns a partition label in $\{0, 1, \ldots, B\kappa-1\}$ to each node in $\bm{X}_{ua}$. The $\text{local\_update}()$ function constructs the graph and computes updated patch features $\bm{X}_{ua}' \in \mathbb{R}^{BN \times C'}$ in parallel across $B\kappa$ partitions, followed by gathering into $\bm{X}'$ via $\text{gather}()$. Notably, (i) $\text{local\_update}$ uses PyTorch Geometric for handling uneven graph sizes efficiently \cite{fey2019fastgraphrepresentationlearning}, and (ii) $\text{scatter}()$ and $\text{gather}()$ leverage PyTorch Scatter for GPU-accelerated operations \cite{pytorch-scatter}. These optimizations further enhance DEGCs training and inference performance.

\subsection{ClusterViG Architecture}
\label{subsec: cvigarch}

\begin{figure*}
    \centering
    \includegraphics[width=\linewidth]{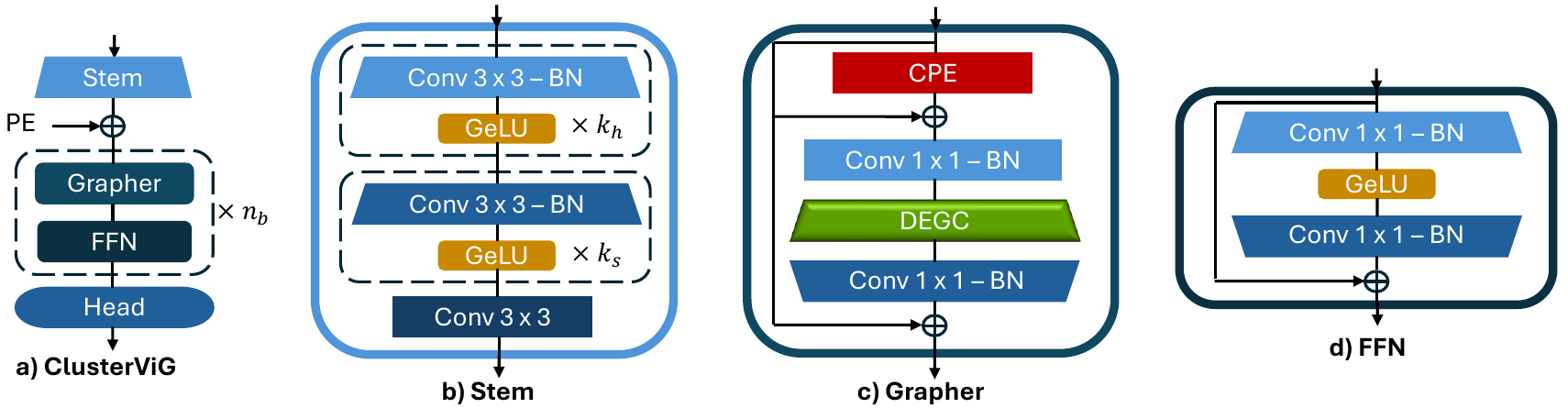}
    \caption{\textbf{ClusterViG} architecture comprised of a pre-processing \textbf{Stem}, a backbone of \textbf{Grapher} and \textbf{FFN} repeated $n_b$ times, and a post-processing \textbf{Head}.}
    \label{fig: clustervig_arch}
\end{figure*}

The ClusterViG architecture and its components are shown in Fig. \ref{fig: clustervig_arch}. It has four main blocks: (i) \emph{Stem} (ii) \emph{Grapher} (iii) \emph{FFN} and (iv) \emph{Head}. The \emph{Grapher} and \emph{FFN} block form a backbone that is repeated $n_b$ times. We now describe each block in detail. (i) \emph{Stem}. The pre-processing convolutional \emph{Stem} (Fig. \ref{fig: clustervig_arch}b) contains $k_h$ $3\times3$ convolution layers with stride $2$ followed by Batch Normalization (BN) and GeLU activation. Each such convolution halves the image resolution input fed to it, with the final output resolution being $H_{\text{iso}} \times W_{\text{iso}}$. The next set of $k_s$ $3\times3$ convolution layers with stride $1$ with BN and GeLU maintain the image resolution. The total $k_h + k_s$ convolution layers increase the image channels from $C_\text{in} = 3$ to $C_\text{iso}$. Finally, a $3\times3$ convolution with stride $1$ is applied that maintains the isotropic image resolution and channels. The output of the \emph{Stem} is an image patch feature grid in $\mathbb{R}^{H_\text{iso} \times W_\text{iso} \times C_\text{iso}}$. (ii) \emph{Grapher}. A single \emph{Grapher} block is shown in Fig. \ref{fig: clustervig_arch}c. Given an input $\bm{X} \in \mathbb{R}^{N_\text{iso} \times C_\text{iso}}$ (tokenized, $N_\text{iso} = H_\text{iso} \times W_\text{iso}$), the output of the single \emph{Grapher} block can be described as below,
\begin{equation}
\label{eq: grapher}
    \bm{Y} = \sigma\rbr{\text{DEGC}((\text{CPE}(\bm{X}) + \bm{X})\bm{W}_1)}\bm{W}_2 + \bm{X}
\end{equation}
In Eq. \ref{eq: grapher}, $\bm{Y} \in \mathbb{R}^{N_{\text{iso}} \times C_\text{iso}}$. This \emph{Grapher} block is based on the \emph{Grapher} block in \cite{vig, greedyvig}. First, it applies Conditional Positional Encoding \cite{cpe} to the input $\bm{X}$, before applying a Fully Connected (FC) layer $\bm{W}_1 \in \mathbb{R}^{C_\text{iso} \times C_\text{iso}}$. The output of $\sigma(\text{DEGC}(\cdot))$ is in $\mathbb{R}^{N_\text{iso} \times 2C_\text{iso}}$ which is converted back to $\mathbb{R}^{N_{\text{iso}} \times C_\text{iso}}$ via FC $\bm{W}_2$. $\sigma$ is the GeLU activation and within DEGC, $\Tilde{C} = C_{\text{iso}}$, where $\Tilde{C}$ is the output channels for the global feature vectors. (iii) \emph{FFN}. The \emph{FFN} (Fig. \ref{fig: clustervig_arch}d) layer receives the input $\bm{Y}$ from \emph{Grapher} and processes it as,
\begin{equation}
    \label{eq: ffn}
    \bm{Z} = \sigma(\bm{Y}\bm{W}'_1)\bm{W}'_2 + \bm{Y}
\end{equation}
Note that $\bm{Z} \in \mathbb{R}^{N_\text{iso} \times C_\text{iso}}$. $\bm{W}_1'$ is an FC that increases the output channels to $4C_\text{iso}$ which is then decreased back to $C_\text{iso}$ via $\bm{W}_2'$. Note that Eq. \ref{eq: grapher} and \ref{eq: ffn} assume tokenized input, and we skip the tokenization and reshaping operations for simplicity. Finally, (iv) \emph{Head} is the standard classification head comprising Average Pooling and FC layers that output class logits.

\section{Experiments}
\label{sec: exp}

We conduct experiments to compare ClusterViG against ViG-based models such as ViG \cite{vig}, ViHGNN \cite{vihgnn}, PVG \cite{pvg}, MobileViG \cite{mobilevig}, and GreedyViG \cite{greedyvig}, along with other efficient vision architectures.  

\subsection{ClusterViG Architectures} Based on the ClusterViG architecture described in Sec. \ref{subsec: cvigarch}, we define four isotropic variants of ClusterViG, with their defining parameters described in Table \ref{tab: clustervig_variants}. Note that $H_\text{iso} = W_\text{iso} = N_\text{iso}^{\frac{1}{2}}$. For all the variants, the neighbors per partition, $K$, increases from $9$ to $18$, linearly over the blocks $n_b$. The total number of partitions, $\kappa$, is set to $4$ for all variants except CViG-B$^{\dag}$, which uses $\kappa = 6$. The default G-GCN used for all the models is G-MRGCN (Table \ref{tab: g-gcn-variant-update-rule}). Note that CViG-B$^{\dag}$ is identical to CViG-B, excluding the fact that it is a $2\times$ resolution variant of ClusterViG. We implement our models with PyTorch 2.2.2 \cite{pytorch}, PyTorch Geometric 2.5.2 \cite{pyg}, and Timm 1.0.9 \cite{timm} libraries.

\begin{table}[]
    \centering
    \caption{Variants of ClusterViG, isotropic}
    \vspace{-5pt}
    \begin{tabular}{cccccc}
         \toprule
         Model & $n_b$ & $C_\text{iso}$ & $N_\text{iso}$ & Params (M) &  GMACs \\
         \midrule
         CViG-Ti &  12   &     192           &       196        &     11.5      & 1.3             \\
         \midrule
         CViG-S  &  16    &    320       &    196   &   28.2    &    4.2   \\
         \midrule 
         CViG-B   &  16   &    640       &    196   &   104.8    &  16.2     \\
         \midrule
         CViG-B$^\dag$   &  16   &    640       &   784    &  105.2     &  62.3     \\
         \bottomrule
    \end{tabular}
    \label{tab: clustervig_variants}
\end{table}

\subsection{CV Benchmarks} We conduct experiments on image classification, object detection, and instance segmentation following prior works \cite{vig, vihgnn, pvg, mobilevig, greedyvig}.

\textbf{Image Classification.} We use 8 NVIDIA A100 GPUs to train our models with an effective batch size of 1024. The models are trained on ImageNet-1K \cite{imagenet} using the AdamW optimizer \cite{adamwoptim} and a learning rate of $2\times 10^{-3}$ with cosine annealing schedule for 300 epochs. The input image resolution for training and testing is $224 \times 224$. Note that all model variants reduce this resolution to $H_{\text{iso}} \times W_{\text{iso}} = 14 \times 14$ for the backbone, excluding CViG-B$^{\dag}$ which has $2\times$ resolution ($28 \times 28$). We perform knowledge distillation using RegNetY-16GF \cite{regnet} following prior works. The data augmentation techniques employed during training are identical to those described in \cite{vig, vihgnn}. The results in Table \ref{Classification_Results} show that ClusterViG outperforms Pyramid ViG (PViG), Pyramid ViHGNN (PViHGNN), PVG, GreedyViG and MobileViG, with similar model parameters and GMACs. We note that ClusterViG reaches this performance with an isotropic architecture. CViG-B and CViG-B$^\dag$ have comparable model parameters to PViG-B, PViHGNN-B, and PVG-B while achieving superior performance. Despite being approximately $3.4\times$ larger, these models deliver substantial performance gains compared to GreedyViG-B. Notably, CViG-S, with its smaller model size, achieves a comparable top-1 accuracy to GreedyViG-B. Input partitioning enables efficient training of models with higher isotropic resolution, such as CViG-B$^{\dag}$, outperforming low-resolution counterparts and related SOTA models. Compared to other efficient architectures, CViG variants outperform them at similar model sizes (e.g., CViG-S surpasses EfficientFormerV2-L with just 7\% more parameters).

\begin{table}[]
\caption{Classification on ImageNet-1k}
\label{Classification_Results}
\vspace{-5pt}
\centering
\resizebox{\columnwidth}{!}{
\begin{tabular}{cccccc}
\toprule
\textbf{Model} & \textbf{Type} & \textbf{Parameters (M)} & \textbf{GMACs} & \textbf{Epochs} & \textbf{Top-1 Accuracy (\%)} \\ 
\midrule
ResNet18 \cite{resnet} & CNN & 11.7 & 1.82 & 300 & 69.7 \\ 
ResNet50 \cite{resnet} & CNN & 25.6 & 4.1 & 300 & 80.4 \\ 
ConvNext-T \cite{liu2022convnet} & CNN & 28.6 & 7.4 & 300 & 82.7 \\ 
\midrule
EfficientFormer-L1 \cite{efficientformer} & CNN-ViT & 12.3 & 1.3 & 300 & 79.2 \\ 
EfficientFormer-L3 \cite{efficientformer} & CNN-ViT & 31.3 & 3.9 & 300 & 82.4 \\ 
EfficientFormer-L7 \cite{efficientformer} & CNN-ViT & 82.1 & 10.2 & 300 & 83.3 \\ 
LeViT-192 \cite{graham2021levit} & CNN-ViT & 10.9 & 0.7 & 1000 & 80.0 \\ 
LeViT-384 \cite{graham2021levit} & CNN-ViT & 39.1 & 2.4 & 1000 & 82.6 \\ 
EfficientFormerV2-S2 \cite{rethinking-vit} & CNN-ViT & 12.6 & 1.3 & 300 & 81.6 \\ 
EfficientFormerV2-L \cite{rethinking-vit} & CNN-ViT & 26.1 & 2.6 & 300 & 83.3 \\ 
\midrule
PVT-Small \cite{wang2021pyramid} & ViT & 24.5 & 3.8 & 300 & 79.8 \\ 
PVT-Large \cite{wang2021pyramid} & ViT & 61.4 & 9.8 & 300 & 81.7 \\ 
DeiT-S \cite{touvron2021training} & ViT & 22.5 & 4.5 & 300 & 81.2 \\ 
Swin-T \cite{liu2022swin} & ViT & 29.0 & 4.5 & 300 & 81.4 \\ 
PoolFormer-s12 \cite{yu2022metaformer} & Pool & 12.0 & 2.0 & 300 & 77.2 \\ 
PoolFormer-s24 \cite{yu2022metaformer} & Pool & 21.0 & 3.6 & 300 & 80.3 \\ 
PoolFormer-s36 \cite{yu2022metaformer} & Pool & 31.0 & 5.2 & 300 & 81.4 \\ 
\midrule
PViHGNN-Ti \cite{vihgnn} & GNN & 12.3 & 2.3 & 300 & 78.9 \\ 
PViHGNN-S \cite{vihgnn} & GNN & 28.5 & 6.3 & 300 & 82.5 \\ 
PViHGNN-B \cite{vihgnn} & GNN & 94.4 & 18.1 & 300 & 83.9 \\ 
\midrule
PViG-Ti \cite{vig} & GNN & 10.7 & 1.7 & 300 & 78.2 \\ 
PViG-S \cite{vig} & GNN & 27.3 & 4.6 & 300 & 82.1 \\ 
PViG-B \cite{vig} & GNN & 92.6 & 16.8 & 300 & 83.7 \\ 
\midrule
PVG-S \cite{pvg} & GNN & 22.0 & 5    & 300 & 83.0 \\
PVG-M \cite{pvg} & GNN & 42.0 & 8.9  & 300 & 83.7 \\
PVG-B \cite{pvg} & GNN & 79.0 & 16.9 & 300 & 84.2 \\ 
\midrule
MobileViG-S \cite{mobilevig} & CNN-GNN & 7.2 & 1.0 & 300 & 78.2 \\ 
MobileViG-M \cite{mobilevig} & CNN-GNN & 14.0 & 1.5 & 300 & 80.6 \\ 
MobileViG-B \cite{mobilevig} & CNN-GNN & 26.7 & 2.8 & 300 & 82.6 \\ 
\midrule
GreedyViG-S \cite{greedyvig} & CNN-GNN & 12.0 & 1.6 & 300 & 81.1 \\ 
GreedyViG-M \cite{greedyvig} & CNN-GNN & 21.9 & 3.2 & 300 & 82.9 \\ 
GreedyViG-B \cite{greedyvig} & CNN-GNN & 30.9 & 5.2 & 300 & 83.9 \\ 
\midrule
\textbf{CViG-Ti (Ours)}          & \textbf{CNN-GNN}               &  \textbf{11.5}  &  \textbf{1.3}  & \textbf{300}   &  \textbf{80.3} \\
\textbf{CViG-S  (Ours)}          & \textbf{CNN-GNN}               &  \textbf{28.2}  &  \textbf{4.2}  & \textbf{300}  & \textbf{83.7} \\
\textbf{CViG-B  (Ours)}          & \textbf{CNN-GNN}               &  \textbf{104.8} &  \textbf{16.2} & \textbf{300}  & 
\textbf{85.6} \\
\textbf{CViG-B$^\dag$ (Ours)}    & \textbf{CNN-GNN}               &  \textbf{105.2} &  \textbf{62.3} & \textbf{300}  &   \textbf{87.2} \\
\bottomrule
\end{tabular}}
\end{table}

\textbf{Object Detection and Instance Segmentation.} We validate the generalization of ClusterViG by using it as a backbone for object detection and instance segmentation downstream tasks on the MS COCO 2017 dataset, using the Mask-RCNN framework \cite{cocotasks}. We use a Feature Pyramid Network (FPN) \cite{fpn} as our neck to extract multi-scale feature maps following prior works \cite{vig, vihgnn, pvg, greedyvig}. CViG-S is used as a backbone for the model, initialized with the ImageNet-1K pre-trained weights. We fine-tune the model with AdamW \cite{adamwoptim} and an initial learning rate of $2\times 10^{-4}$, for 12 epochs with a resolution of $1333 \times 800$ on 8 NVIDIA RTX 6000 Ada GPUs, similar to previous works \cite{vig, vihgnn, pvg, greedyvig}. Table \ref{Object_Detection_Segmentation_Results} shows that CViG-S outperforms related (ViG-based and efficient vision) models with similar or fewer model parameters. This highlights the effectiveness of the DEGC \emph{module} within ClusterViG in seamlessly integrating local and global feature learning.

\begin{table}[]
\caption{Object detection and instance segmentation on MS COCO 2017 (\textbf{Parameters} are the backbone model parameters)}
\label{Object_Detection_Segmentation_Results}
\vspace{-5pt}
\centering
\resizebox{\columnwidth}{!}{
\begin{tabular}{cccccccccc}
\toprule
{\textbf{Backbone}} & {\textbf{Parameters (M)}} & AP$^\text{box}$ & AP$^\text{box}_{50}$ & AP$^\text{box}_{75}$ & AP$^\text{mask}$ & AP$^\text{mask}_{50}$ & AP$^\text{mask}_{75}$ \
\\ \midrule

ResNet50 \cite{resnet}            & 25.5 & 38.0 & 58.6 & 41.4 & 34.4 & 55.1 & 36.7     \\ \midrule
EfficientFormer-L3 \cite{efficientformer}    & 31.3 & 41.4 & 63.9 & 44.7 & 38.1 & 61.0 & 40.4     \\ \midrule
EfficientFormer-L7 \cite{efficientformer}    & 82.1 &  42.6 & 65.1 & 46.1 & 39.0 & 62.2 & 41.7        \\ \midrule
EfficientFormerV2-L \cite{rethinking-vit}      & 26.1 & 44.7 & 66.3 & 48.8 & 40.4 & 63.5 & 43.2    \\ \midrule
PoolFormer-S24 \cite{yu2022metaformer}          & 21.0 & 40.1 & 62.2 & 43.4 & 37.0 & 59.1 & 39.6     \\ \midrule
FastViT-SA36 \cite{FastViT}                   & 30.4 & 43.8 & 65.1 & 47.9 & 39.4 & 62.0 & 42.3  \\ \midrule
Pyramid ViG-S \cite{vig}       & 27.3 & 42.6 & 65.2 & 46.0 & 39.4 & 62.4 & 41.6    \\ \midrule
PVG-S & 22.0 & 43.9 & 66.3 & 48.0 & 39.8 & 62.8 & 42.4 \\ \midrule
Pyramid ViHGNN-S \cite{vihgnn}           & 28.5 & 43.1 & 66.0 & 46.5 & 39.6 & 63.0 & 42.3  \\ \midrule
PVT-Small \cite{wang2021pyramid}         & 24.5 & 40.4 & 62.9 & 43.8 & 37.8 & 60.1 & 40.3   \\ \midrule
MobileViG-B \cite{mobilevig}   & 26.7 & 42.0 & 64.3 & 46.0 & 38.9 & 61.4 & 41.6   \\ \midrule
GreedyViG-B \cite{greedyvig}    & 30.9 & 46.3 & 68.4 & 51.3 & 42.1 & 65.5 & 45.4     \\ \midrule
\textbf{CViG-S (Ours)}   &  \textbf{28.2}     &  \textbf{47.4}    &  \textbf{68.1}     &   \textbf{52.0}    &  \textbf{43.4}     &   \textbf{67.2}    &  \textbf{47.5}    
\\ \bottomrule
\end{tabular}}
\end{table}

\begin{figure*}[]
    \centering
    \begin{subfigure}{0.32\textwidth}
        \centering
        \includegraphics[width=\linewidth]{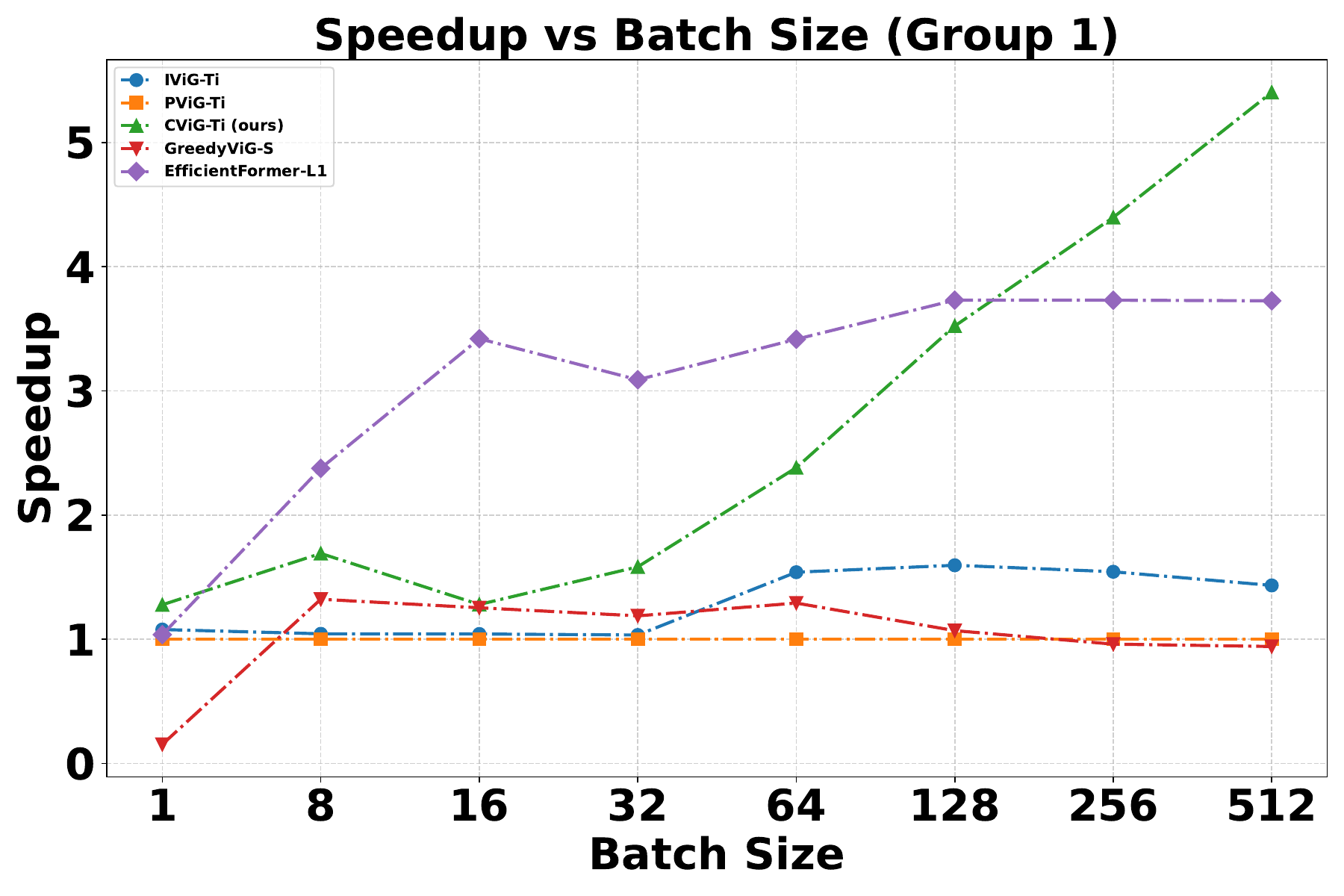}
        \caption{Group 1}
        \label{fig: group1}
    \end{subfigure}
    \hfill
    \begin{subfigure}{0.32\textwidth}
        \centering
        \includegraphics[width=\linewidth]{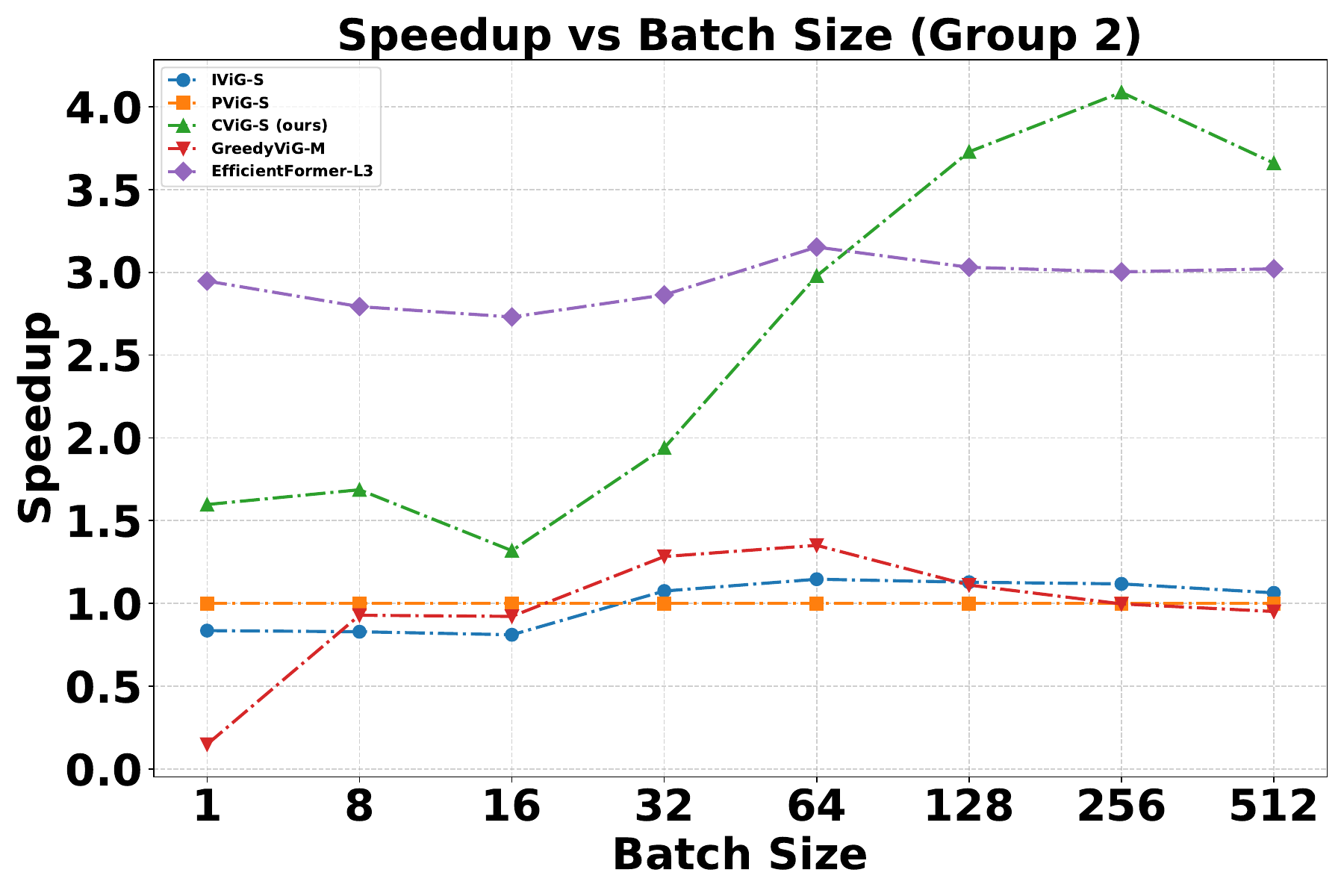}
        \caption{Group 2}
        \label{fig: group2}
    \end{subfigure}
    \hfill
    \begin{subfigure}{0.32\textwidth}
        \centering
        \includegraphics[width=\linewidth]{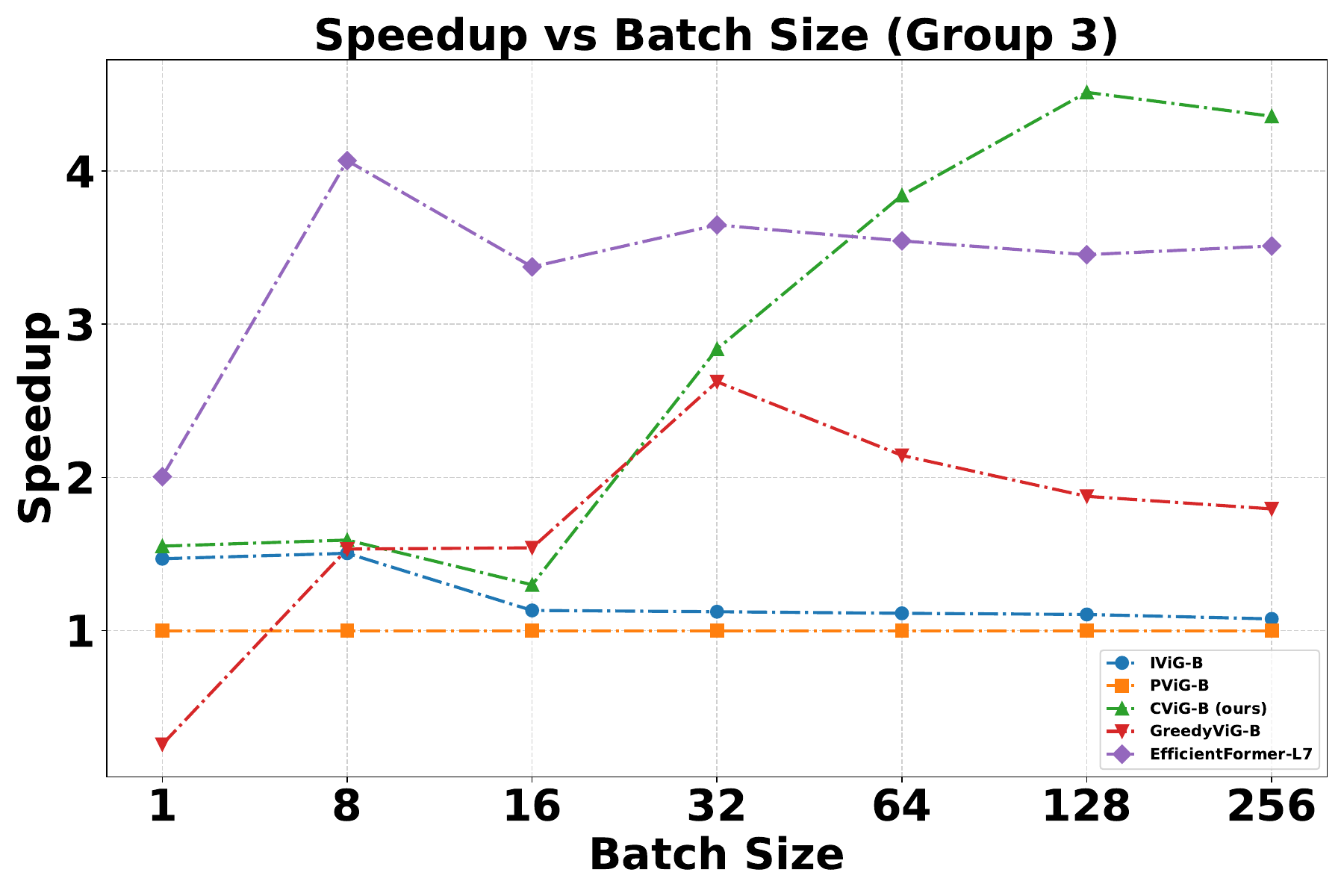}
        \caption{Group 3}
        \label{fig: group3}
    \end{subfigure}

    \caption{Comparing the inference performance of models in three groups.}
    \label{fig:inf_perf}
\end{figure*}

\subsection{Inference Performance} 

We compare the end-to-end inference performance of ClusterViG against the following models: Isotropic ViG (IViG) \cite{vig}, PViG \cite{vig}, GreedyViG \cite{greedyvig}, and EfficientFormer \cite{efficientformer}. We categorize the model variants into three groups based on similar model sizes to ensure a fair comparison. Figure \ref{fig:inf_perf} illustrates the inference speedup as a function of batch size for the model variants within each group. The speedup is defined as the ratio of the baseline model's latency to the latency of the compared model. PViG is the baseline at each batch size, and the speedup is computed relative to its latency for each model. Latency measurements are performed on a single NVIDIA RTX 6000 Ada GPU and averaged over 50 runs. ClusterViG, across all variants, outperforms other ViG-based models while performing as well as EfficientFormer for smaller batch sizes. Notably, for larger batch sizes, ClusterViG performs better than EfficientFormer. Despite being $3.4\times$ larger than GreedyViG-B, CViG-B significantly outperforms it at larger batch sizes. This further underscores the scalability of ClusterViG. The throughput for ViG-based models, computed for different batch sizes, is shown in \ref{tab: throughput}. ClusterViG variants show a consistent trend of increase in throughput with an increase in batch size. We further note that, despite CViG-B$^{\dagger}$ operating at $2\times$ the resolution of PViG-B, it achieves comparable throughput without encountering an out-of-memory error at a batch size of $b = 512$.

\begin{table}[]
\centering
\caption{Throughput measured at batch size $b$ (ViG-based models), note that (-) refers to out-of-memory (OOM) error}
\label{tab: throughput}
\resizebox{\columnwidth}{!}{
\begin{tabular}{ccccccccc}
\toprule
\multirow{2}{*}{\textbf{Model}} & \multicolumn{8}{c}{\textbf{Throughput} (images per second)} \\ \cmidrule(lr){2-9}
      & $b=1$ & $b=8$ & $b=16$ & $b=32$ & $b=64$ & $b=128$ & $b=256$ & $b=512$ \\ 
\midrule
PViG-Ti                  & 28.4 & 241.7 & 480.2 & 966.1 & 1063.6 & 993.0 & 956.9 & 913.9      \\ 
PViG-S                   & 28.2 & 233.9 & 476.4 & 622.5 & 580.4& 552.6& 526.2& 507.3       \\
PViG-B                   & 15.3 & 124.4 & 245.1 & 236.0 & 224.9 & 213.0 & 199.4 & -    \\
\midrule
GreedyViG-S              & 4.3& 319.5& 601.6& 1147.6& 1372.9& 1061.4& 918.7& 860.2      \\
GreedyViG-M              & 4.2 & 217.3 & 439.2 & 798.9 & 784.2 & 613.9 & 524.7 & 482.6     \\
GreedyViG-B               &  3.9 & 190.7 & 377.8 & 619.5 & 482.3 & 400.0 & 358.0 & 330.2      \\
\midrule
\textbf{CViG-Ti (Ours) }                 & 36.3 & 408.7 & 614.4 & 1529.1 & 2532.7 & 3497.6 & 4207.0 & 4938.8      \\
\textbf{CViG-S (Ours) }                  & 45.1 & 394.7 & 628.3 & 1207.5 & 1728.9 & 2060.4 & 2151.0 & 1857.0     \\
\textbf{CViG-B (Ours) }                 & 23.8 & 198.0 & 318.5 & 670.0 & 864.2 & 961.9 & 869.2 & 758.5
     \\
\textbf{CViG-B$^{\dagger}$ (Ours) }      & 17.9 & 91.4 & 144.4 & 206.9 & 191.4 & 159.9 & 151.4 & 156.2
    \\
\bottomrule
\end{tabular}}

\end{table}

\section{Conclusion And Future Work}
\label{sec: concl}

In this paper, we propose a novel method, DEGC, to build a new class of ViG models called ClusterViG. DEGC reduces the computational expense of image graph construction by partitioning the image patch feature grid into several partitions and performing graph construction in parallel within each partition. It leverages global partition-level features learned via GATv2 to enhance graph representation learning. These global features are then incorporated using the proposed G-GCN, enabling globally-aware local feature learning in parallel across all partitions. We use DEGC to develop isotropic ClusterViG variants, including a high-resolution variant. Extensive experiments on benchmark computer vision tasks demonstrate ClusterViG's superior performance compared to state-of-the-art ViG-based and efficient vision models. ClusterViG is significantly faster than its counterparts and exhibits exceptional scalability in inference performance, making it both fast and flexible. Future work will explore how different partitioning strategies impact ClusterViG's performance, as well as more efficient ways to partition the input image. We will also investigate the effect of using different G-GCN variants, varying the number of partitions, and adjusting the number of neighbors per partition. Methods to identify optimal parameters for these configurations will be explored. Finally, accelerating ClusterViG on various hardware platforms for edge applications presents an exciting direction for future research.

\section*{Acknowledgement}
This work is supported by the DEVCOM Army Research Lab (ARL) under grants W911NF2220159 and W911NF2320186 and the National Science Foundation (NSF) under grant SaTC-2104264.
\textbf{Distribution Statement A}: Approved for public release. Distribution is unlimited.

\bibliographystyle{IEEEtran}
\bibliography{ref}

\begin{thebibliography}{10}
\providecommand{\url}[1]{#1}
\csname url@samestyle\endcsname
\providecommand{\newblock}{\relax}
\providecommand{\bibinfo}[2]{#2}
\providecommand{\BIBentrySTDinterwordspacing}{\spaceskip=0pt\relax}
\providecommand{\BIBentryALTinterwordstretchfactor}{4}
\providecommand{\BIBentryALTinterwordspacing}{\spaceskip=\fontdimen2\font plus
\BIBentryALTinterwordstretchfactor\fontdimen3\font minus \fontdimen4\font\relax}
\providecommand{\BIBforeignlanguage}[2]{{%
\expandafter\ifx\csname l@#1\endcsname\relax
\typeout{** WARNING: IEEEtran.bst: No hyphenation pattern has been}%
\typeout{** loaded for the language `#1'. Using the pattern for}%
\typeout{** the default language instead.}%
\else
\language=\csname l@#1\endcsname
\fi
#2}}
\providecommand{\BIBdecl}{\relax}
\BIBdecl

\bibitem{qiu2018deepinf}
J.~Qiu, J.~Tang, H.~Ma, Y.~Dong, K.~Wang, and J.~Tang, ``Deepinf: Social influence prediction with deep learning,'' in \emph{Proceedings of the 24th ACM SIGKDD international conference on knowledge discovery \& data mining}, 2018, pp. 2110--2119.

\bibitem{zhang2018link}
M.~Zhang and Y.~Chen, ``Link prediction based on graph neural networks,'' \emph{Advances in neural information processing systems}, vol.~31, 2018.

\bibitem{zhou-etal-2019-gear}
\BIBentryALTinterwordspacing
J.~Zhou, X.~Han, C.~Yang, Z.~Liu, L.~Wang, C.~Li, and M.~Sun, ``{GEAR}: Graph-based evidence aggregating and reasoning for fact verification,'' in \emph{Proceedings of the 57th Annual Meeting of the Association for Computational Linguistics}, A.~Korhonen, D.~Traum, and L.~M{\`a}rquez, Eds.\hskip 1em plus 0.5em minus 0.4em\relax Florence, Italy: Association for Computational Linguistics, Jul. 2019, pp. 892--901. [Online]. Available: \url{https://aclanthology.org/P19-1085}
\BIBentrySTDinterwordspacing

\bibitem{qiu-etal-2019-dynamically}
\BIBentryALTinterwordspacing
L.~Qiu, Y.~Xiao, Y.~Qu, H.~Zhou, L.~Li, W.~Zhang, and Y.~Yu, ``Dynamically fused graph network for multi-hop reasoning,'' in \emph{Proceedings of the 57th Annual Meeting of the Association for Computational Linguistics}, A.~Korhonen, D.~Traum, and L.~M{\`a}rquez, Eds.\hskip 1em plus 0.5em minus 0.4em\relax Florence, Italy: Association for Computational Linguistics, Jul. 2019, pp. 6140--6150. [Online]. Available: \url{https://aclanthology.org/P19-1617}
\BIBentrySTDinterwordspacing

\bibitem{wang2019kgat}
X.~Wang, X.~He, Y.~Cao, M.~Liu, and T.-S. Chua, ``Kgat: Knowledge graph attention network for recommendation,'' in \emph{Proceedings of the 25th ACM SIGKDD international conference on knowledge discovery \& data mining}, 2019, pp. 950--958.

\bibitem{wang2020multi}
X.~Wang, R.~Wang, C.~Shi, G.~Song, and Q.~Li, ``Multi-component graph convolutional collaborative filtering,'' in \emph{Proceedings of the AAAI conference on artificial intelligence}, vol.~34, no.~04, 2020, pp. 6267--6274.

\bibitem{cho2018three}
H.~Cho and I.~S. Choi, ``Three-dimensionally embedded graph convolutional network (3dgcn) for molecule interpretation,'' \emph{arXiv preprint arXiv:1811.09794}, 2018.

\bibitem{hoshen2017vain}
Y.~Hoshen, ``Vain: Attentional multi-agent predictive modeling,'' \emph{Advances in neural information processing systems}, vol.~30, 2017.

\bibitem{pmlr-v80-sanchez-gonzalez18a}
A.~Sanchez-Gonzalez, N.~Heess, J.~T. Springenberg, J.~Merel, M.~Riedmiller, R.~Hadsell, and P.~Battaglia, ``Graph networks as learnable physics engines for inference and control,'' in \emph{Proceedings of the 35th International Conference on Machine Learning}, ser. Proceedings of Machine Learning Research, J.~Dy and A.~Krause, Eds., vol.~80.\hskip 1em plus 0.5em minus 0.4em\relax PMLR, 10--15 Jul 2018, pp. 4470--4479.

\bibitem{ktena2017distance}
S.~I. Ktena, S.~Parisot, E.~Ferrante, M.~Rajchl, M.~Lee, B.~Glocker, and D.~Rueckert, ``Distance metric learning using graph convolutional networks: Application to functional brain networks,'' in \emph{Medical Image Computing and Computer Assisted Intervention- MICCAI 2017: 20th International Conference, Quebec City, QC, Canada, September 11-13, 2017, Proceedings, Part I 20}.\hskip 1em plus 0.5em minus 0.4em\relax Springer, 2017, pp. 469--477.

\bibitem{zhou2022graph}
Y.~Zhou, H.~Zheng, X.~Huang, S.~Hao, D.~Li, and J.~Zhao, ``Graph neural networks: Taxonomy, advances, and trends,'' \emph{ACM Transactions on Intelligent Systems and Technology (TIST)}, vol.~13, no.~1, pp. 1--54, 2022.

\bibitem{alexnet}
A.~Krizhevsky, I.~Sutskever, and G.~E. Hinton, ``Imagenet classification with deep convolutional neural networks,'' in \emph{Advances in Neural Information Processing Systems}, F.~Pereira, C.~Burges, L.~Bottou, and K.~Weinberger, Eds., vol.~25.\hskip 1em plus 0.5em minus 0.4em\relax Curran Associates, Inc., 2012.

\bibitem{efficientnet}
M.~Tan and Q.~Le, ``Efficientnet: Rethinking model scaling for convolutional neural networks,'' in \emph{Proceedings of the 36th International Conference on Machine Learning}, ser. Proceedings of Machine Learning Research, K.~Chaudhuri and R.~Salakhutdinov, Eds., vol.~97.\hskip 1em plus 0.5em minus 0.4em\relax PMLR, 09--15 Jun 2019, pp. 6105--6114.

\bibitem{resnet}
K.~He, X.~Zhang, S.~Ren, and J.~Sun, ``Deep residual learning for image recognition,'' in \emph{Proceedings of the IEEE conference on computer vision and pattern recognition}, 2016, pp. 770--778.

\bibitem{Cai_2018_CVPR}
Z.~Cai and N.~Vasconcelos, ``Cascade r-cnn: Delving into high quality object detection,'' in \emph{Proceedings of the IEEE Conference on Computer Vision and Pattern Recognition (CVPR)}, June 2018.

\bibitem{fcn_semantic}
J.~Long, E.~Shelhamer, and T.~Darrell, ``Fully convolutional networks for semantic segmentation,'' in \emph{Proceedings of the IEEE conference on computer vision and pattern recognition}, 2015, pp. 3431--3440.

\bibitem{instance_seg}
X.~Wang, R.~Zhang, T.~Kong, L.~Li, and C.~Shen, ``Solov2: Dynamic and fast instance segmentation,'' in \emph{Advances in Neural Information Processing Systems}, H.~Larochelle, M.~Ranzato, R.~Hadsell, M.~Balcan, and H.~Lin, Eds., vol.~33.\hskip 1em plus 0.5em minus 0.4em\relax Curran Associates, Inc., 2020, pp. 17\,721--17\,732.

\bibitem{tolstikhin2021mlp}
I.~O. Tolstikhin, N.~Houlsby, A.~Kolesnikov, L.~Beyer, X.~Zhai, T.~Unterthiner, J.~Yung, A.~Steiner, D.~Keysers, J.~Uszkoreit \emph{et~al.}, ``Mlp-mixer: An all-mlp architecture for vision,'' \emph{Advances in neural information processing systems}, vol.~34, pp. 24\,261--24\,272, 2021.

\bibitem{touvron2022resmlp}
H.~Touvron, P.~Bojanowski, M.~Caron, M.~Cord, A.~El-Nouby, E.~Grave, G.~Izacard, A.~Joulin, G.~Synnaeve, J.~Verbeek \emph{et~al.}, ``Resmlp: Feedforward networks for image classification with data-efficient training,'' \emph{IEEE transactions on pattern analysis and machine intelligence}, vol.~45, no.~4, pp. 5314--5321, 2022.

\bibitem{vaswani2017attention}
A.~Vaswani, ``Attention is all you need,'' \emph{Advances in Neural Information Processing Systems}, 2017.

\bibitem{dosovitskiy2021an}
\BIBentryALTinterwordspacing
A.~Dosovitskiy, L.~Beyer, A.~Kolesnikov, D.~Weissenborn, X.~Zhai, T.~Unterthiner, M.~Dehghani, M.~Minderer, G.~Heigold, S.~Gelly, J.~Uszkoreit, and N.~Houlsby, ``An image is worth 16x16 words: Transformers for image recognition at scale,'' in \emph{International Conference on Learning Representations}, 2021. [Online]. Available: \url{https://openreview.net/forum?id=YicbFdNTTy}
\BIBentrySTDinterwordspacing

\bibitem{li2022uniformer}
K.~Li, Y.~Wang, P.~Gao, G.~Song, Y.~Liu, H.~Li, and Y.~Qiao, ``Uniformer: Unified transformer for efficient spatiotemporal representation learning,'' \emph{arXiv preprint arXiv:2201.04676}, 2022.

\bibitem{liu2022swin}
Z.~Liu, H.~Hu, Y.~Lin, Z.~Yao, Z.~Xie, Y.~Wei, J.~Ning, Y.~Cao, Z.~Zhang, L.~Dong \emph{et~al.}, ``Swin transformer v2: Scaling up capacity and resolution,'' in \emph{Proceedings of the IEEE/CVF conference on computer vision and pattern recognition}, 2022, pp. 12\,009--12\,019.

\bibitem{carion2020end}
N.~Carion, F.~Massa, G.~Synnaeve, N.~Usunier, A.~Kirillov, and S.~Zagoruyko, ``End-to-end object detection with transformers,'' in \emph{European conference on computer vision}.\hskip 1em plus 0.5em minus 0.4em\relax Springer, 2020, pp. 213--229.

\bibitem{chen2021topological}
K.~Chen, J.~K. Chen, J.~Chuang, M.~V{\'a}zquez, and S.~Savarese, ``Topological planning with transformers for vision-and-language navigation,'' in \emph{Proceedings of the IEEE/CVF Conference on Computer Vision and Pattern Recognition}, 2021, pp. 11\,276--11\,286.

\bibitem{greedyvig}
M.~Munir, W.~Avery, M.~M. Rahman, and R.~Marculescu, ``Greedyvig: Dynamic axial graph construction for efficient vision gnns,'' in \emph{Proceedings of the IEEE/CVF Conference on Computer Vision and Pattern Recognition}, 2024, pp. 6118--6127.

\bibitem{vig}
K.~Han, Y.~Wang, J.~Guo, Y.~Tang, and E.~Wu, ``Vision gnn: An image is worth graph of nodes,'' \emph{Advances in neural information processing systems}, vol.~35, pp. 8291--8303, 2022.

\bibitem{vihgnn}
Y.~Han, P.~Wang, S.~Kundu, Y.~Ding, and Z.~Wang, ``Vision hgnn: An image is more than a graph of nodes,'' in \emph{Proceedings of the IEEE/CVF International Conference on Computer Vision}, 2023, pp. 19\,878--19\,888.

\bibitem{pvg}
J.~Wu, J.~Li, J.~Zhang, B.~Zhang, M.~Chi, Y.~Wang, and C.~Wang, ``Pvg: Progressive vision graph for vision recognition,'' in \emph{Proceedings of the 31st ACM International Conference on Multimedia}, 2023, pp. 2477--2486.

\bibitem{mobilevig}
M.~Munir, W.~Avery, and R.~Marculescu, ``Mobilevig: Graph-based sparse attention for mobile vision applications,'' in \emph{Proceedings of the IEEE/CVF Conference on Computer Vision and Pattern Recognition}, 2023, pp. 2211--2219.

\bibitem{scalemobilevig}
W.~Avery, M.~Munir, and R.~Marculescu, ``Scaling graph convolutions for mobile vision,'' in \emph{Proceedings of the IEEE/CVF Conference on Computer Vision and Pattern Recognition}, 2024, pp. 5857--5865.

\bibitem{imagenet}
J.~Deng, W.~Dong, R.~Socher, L.-J. Li, K.~Li, and L.~Fei-Fei, ``Imagenet: A large-scale hierarchical image database,'' in \emph{2009 IEEE Conference on Computer Vision and Pattern Recognition}, 2009, pp. 248--255.

\bibitem{cocotasks}
T.-Y. Lin, M.~Maire, S.~Belongie, J.~Hays, P.~Perona, D.~Ramanan, P.~Doll{\'a}r, and C.~L. Zitnick, ``Microsoft coco: Common objects in context,'' in \emph{Computer Vision--ECCV 2014: 13th European Conference, Zurich, Switzerland, September 6-12, 2014, Proceedings, Part V 13}.\hskip 1em plus 0.5em minus 0.4em\relax Springer, 2014, pp. 740--755.

\bibitem{denseconnectedcnn}
G.~Huang, Z.~Liu, L.~Van Der~Maaten, and K.~Q. Weinberger, ``Densely connected convolutional networks,'' in \emph{Proceedings of the IEEE conference on computer vision and pattern recognition}, 2017, pp. 4700--4708.

\bibitem{vit-survey}
S.~Khan, M.~Naseer, M.~Hayat, S.~W. Zamir, F.~S. Khan, and M.~Shah, ``Transformers in vision: A survey,'' \emph{ACM computing surveys (CSUR)}, vol.~54, no. 10s, pp. 1--41, 2022.

\bibitem{mobilenet}
A.~G. Howard, ``Mobilenets: Efficient convolutional neural networks for mobile vision applications,'' \emph{arXiv preprint arXiv:1704.04861}, 2017.

\bibitem{mobilenetv2}
M.~Sandler, A.~Howard, M.~Zhu, A.~Zhmoginov, and L.-C. Chen, ``Mobilenetv2: Inverted residuals and linear bottlenecks,'' in \emph{Proceedings of the IEEE conference on computer vision and pattern recognition}, 2018, pp. 4510--4520.

\bibitem{efficientnetv2}
M.~Tan and Q.~Le, ``Efficientnetv2: Smaller models and faster training,'' in \emph{International conference on machine learning}.\hskip 1em plus 0.5em minus 0.4em\relax PMLR, 2021, pp. 10\,096--10\,106.

\bibitem{fast-vit-hilo}
Z.~Pan, J.~Cai, and B.~Zhuang, ``Fast vision transformers with hilo attention,'' \emph{Advances in Neural Information Processing Systems}, vol.~35, pp. 14\,541--14\,554, 2022.

\bibitem{A-vit}
H.~Yin, A.~Vahdat, J.~M. Alvarez, A.~Mallya, J.~Kautz, and P.~Molchanov, ``A-vit: Adaptive tokens for efficient vision transformer,'' in \emph{Proceedings of the IEEE/CVF conference on computer vision and pattern recognition}, 2022, pp. 10\,809--10\,818.

\bibitem{efficientvit}
X.~Liu, H.~Peng, N.~Zheng, Y.~Yang, H.~Hu, and Y.~Yuan, ``Efficientvit: Memory efficient vision transformer with cascaded group attention,'' in \emph{Proceedings of the IEEE/CVF Conference on Computer Vision and Pattern Recognition}, 2023, pp. 14\,420--14\,430.

\bibitem{efficientformer}
Y.~Li, G.~Yuan, Y.~Wen, J.~Hu, G.~Evangelidis, S.~Tulyakov, Y.~Wang, and J.~Ren, ``Efficientformer: Vision transformers at mobilenet speed,'' \emph{Advances in Neural Information Processing Systems}, vol.~35, pp. 12\,934--12\,949, 2022.

\bibitem{rethinking-vit}
Y.~Li, J.~Hu, Y.~Wen, G.~Evangelidis, K.~Salahi, Y.~Wang, S.~Tulyakov, and J.~Ren, ``Rethinking vision transformers for mobilenet size and speed,'' in \emph{Proceedings of the IEEE/CVF International Conference on Computer Vision}, 2023, pp. 16\,889--16\,900.

\bibitem{separable-vit-self-att}
S.~Mehta and M.~Rastegari, ``Separable self-attention for mobile vision transformers,'' \emph{arXiv preprint arXiv:2206.02680}, 2022.

\bibitem{fast-vit-struct-reparam}
P.~K.~A. Vasu, J.~Gabriel, J.~Zhu, O.~Tuzel, and A.~Ranjan, ``Fastvit: A fast hybrid vision transformer using structural reparameterization,'' in \emph{Proceedings of the IEEE/CVF International Conference on Computer Vision}, 2023, pp. 5785--5795.

\bibitem{cyclemlp}
\BIBentryALTinterwordspacing
S.~Chen, E.~Xie, C.~GE, R.~Chen, D.~Liang, and P.~Luo, ``Cycle{MLP}: A {MLP}-like architecture for dense prediction,'' in \emph{International Conference on Learning Representations}, 2022. [Online]. Available: \url{https://openreview.net/forum?id=NMEceG4v69Y}
\BIBentrySTDinterwordspacing

\bibitem{phase-vision-mlp}
Y.~Tang, K.~Han, J.~Guo, C.~Xu, Y.~Li, C.~Xu, and Y.~Wang, ``An image patch is a wave: Phase-aware vision mlp,'' in \emph{Proceedings of the IEEE/CVF conference on computer vision and pattern recognition}, 2022, pp. 10\,935--10\,944.

\bibitem{graphsage}
W.~Hamilton, Z.~Ying, and J.~Leskovec, ``Inductive representation learning on large graphs,'' \emph{Advances in neural information processing systems}, vol.~30, 2017.

\bibitem{gcn-kipf}
T.~N. Kipf and M.~Welling, ``Semi-supervised classification with graph convolutional networks,'' \emph{arXiv preprint arXiv:1609.02907}, 2016.

\bibitem{gnn-survey-2020}
J.~Zhou, G.~Cui, S.~Hu, Z.~Zhang, C.~Yang, Z.~Liu, L.~Wang, C.~Li, and M.~Sun, ``Graph neural networks: A review of methods and applications,'' \emph{AI open}, vol.~1, pp. 57--81, 2020.

\bibitem{edgeconv}
Y.~Wang, Y.~Sun, Z.~Liu, S.~E. Sarma, M.~M. Bronstein, and J.~M. Solomon, ``Dynamic graph cnn for learning on point clouds,'' \emph{ACM Transactions on Graphics (tog)}, vol.~38, no.~5, pp. 1--12, 2019.

\bibitem{mrgcn}
G.~Li, M.~Muller, A.~Thabet, and B.~Ghanem, ``Deepgcns: Can gcns go as deep as cnns?'' in \emph{Proceedings of the IEEE/CVF international conference on computer vision}, 2019, pp. 9267--9276.

\bibitem{geomgcn}
S.~Srivastava and G.~Sharma, ``Exploiting local geometry for feature and graph construction for better 3d point cloud processing with graph neural networks,'' in \emph{2021 IEEE INternational conference on robotics and automation (ICRA)}.\hskip 1em plus 0.5em minus 0.4em\relax IEEE, 2021, pp. 12\,903--12\,909.

\bibitem{agcn}
S.~Kim and D.~C. Alexander, ``Agcn: Adversarial graph convolutional network for 3d point cloud segmentation,'' in \emph{32nd British Machine Vision Conference, BMVC 2021}.\hskip 1em plus 0.5em minus 0.4em\relax The British Machine Vision Association (BMVA), 2021, pp. 1--13.

\bibitem{scenegraphgeneration}
H.~Li, G.~Zhu, L.~Zhang, Y.~Jiang, Y.~Dang, H.~Hou, P.~Shen, X.~Zhao, S.~A.~A. Shah, and M.~Bennamoun, ``Scene graph generation: A comprehensive survey,'' \emph{Neurocomputing}, vol. 566, p. 127052, 2024.

\bibitem{gnn-human-action}
M.~Li, S.~Chen, X.~Chen, Y.~Zhang, Y.~Wang, and Q.~Tian, ``Symbiotic graph neural networks for 3d skeleton-based human action recognition and motion prediction,'' \emph{IEEE transactions on pattern analysis and machine intelligence}, vol.~44, no.~6, pp. 3316--3333, 2021.

\bibitem{gnn-in-vision-survey}
C.~Chen, Y.~Wu, Q.~Dai, H.-Y. Zhou, M.~Xu, S.~Yang, X.~Han, and Y.~Yu, ``A survey on graph neural networks and graph transformers in computer vision: A task-oriented perspective,'' \emph{IEEE Transactions on Pattern Analysis and Machine Intelligence}, 2024.

\bibitem{mpgnn}
\BIBentryALTinterwordspacing
J.~Gilmer, S.~S. Schoenholz, P.~F. Riley, O.~Vinyals, and G.~E. Dahl, ``Neural message passing for quantum chemistry,'' in \emph{Proceedings of the 34th International Conference on Machine Learning}, ser. Proceedings of Machine Learning Research, D.~Precup and Y.~W. Teh, Eds., vol.~70.\hskip 1em plus 0.5em minus 0.4em\relax PMLR, 06--11 Aug 2017, pp. 1263--1272. [Online]. Available: \url{https://proceedings.mlr.press/v70/gilmer17a.html}
\BIBentrySTDinterwordspacing

\bibitem{gin}
K.~Xu, W.~Hu, J.~Leskovec, and S.~Jegelka, ``How powerful are graph neural networks?'' \emph{arXiv preprint arXiv:1810.00826}, 2018.

\bibitem{gatv2}
S.~Brody, U.~Alon, and E.~Yahav, ``How attentive are graph attention networks?'' \emph{arXiv preprint arXiv:2105.14491}, 2021.

\bibitem{zhou2020effect_kmeans}
K.~Zhou and S.~Yang, ``Effect of cluster size distribution on clustering: a comparative study of k-means and fuzzy c-means clustering,'' \emph{Pattern Analysis and Applications}, vol.~23, no.~1, pp. 455--466, 2020.

\bibitem{fey2019fastgraphrepresentationlearning}
\BIBentryALTinterwordspacing
M.~Fey and J.~E. Lenssen, ``Fast graph representation learning with pytorch geometric,'' 2019. [Online]. Available: \url{https://arxiv.org/abs/1903.02428}
\BIBentrySTDinterwordspacing

\bibitem{pytorch-scatter}
{PyTorch Geometric Team}, ``Pytorch scatter documentation,'' \url{https://pytorch-scatter.readthedocs.io/en/latest/}, 2023, accessed: 2025-01-12.

\bibitem{cpe}
X.~Chu, Z.~Tian, B.~Zhang, X.~Wang, and C.~Shen, ``Conditional positional encodings for vision transformers,'' \emph{arXiv preprint arXiv:2102.10882}, 2021.

\bibitem{pytorch}
\BIBentryALTinterwordspacing
P.~Team, ``Pytorch,'' 2016, accessed: January 14, 2025. [Online]. Available: \url{https://pytorch.org/}
\BIBentrySTDinterwordspacing

\bibitem{pyg}
\BIBentryALTinterwordspacing
P.~G. Team, ``Pytorch geometric,'' 2018, accessed: January 14, 2025. [Online]. Available: \url{https://pytorch-geometric.readthedocs.io/en/latest/}
\BIBentrySTDinterwordspacing

\bibitem{timm}
\BIBentryALTinterwordspacing
R.~Wightman, ``Pytorch image models (timm),'' 2019, accessed: January 14, 2025. [Online]. Available: \url{https://timm.fast.ai/}
\BIBentrySTDinterwordspacing

\bibitem{adamwoptim}
I.~Loshchilov, ``Decoupled weight decay regularization,'' \emph{arXiv preprint arXiv:1711.05101}, 2017.

\bibitem{regnet}
I.~Radosavovic, R.~P. Kosaraju, R.~Girshick, K.~He, and P.~Doll{\'a}r, ``Designing network design spaces,'' in \emph{Proceedings of the IEEE/CVF conference on computer vision and pattern recognition}, 2020, pp. 10\,428--10\,436.

\bibitem{liu2022convnet}
Z.~Liu, H.~Mao, C.-Y. Wu, C.~Feichtenhofer, T.~Darrell, and S.~Xie, ``A convnet for the 2020s,'' in \emph{Proceedings of the IEEE/CVF conference on computer vision and pattern recognition}, 2022, pp. 11\,976--11\,986.

\bibitem{graham2021levit}
B.~Graham, A.~El-Nouby, H.~Touvron, P.~Stock, A.~Joulin, H.~J{\'e}gou, and M.~Douze, ``Levit: a vision transformer in convnet's clothing for faster inference,'' in \emph{Proceedings of the IEEE/CVF international conference on computer vision}, 2021, pp. 12\,259--12\,269.

\bibitem{wang2021pyramid}
W.~Wang, E.~Xie, X.~Li, D.-P. Fan, K.~Song, D.~Liang, T.~Lu, P.~Luo, and L.~Shao, ``Pyramid vision transformer: A versatile backbone for dense prediction without convolutions,'' in \emph{Proceedings of the IEEE/CVF international conference on computer vision}, 2021, pp. 568--578.

\bibitem{touvron2021training}
H.~Touvron, M.~Cord, M.~Douze, F.~Massa, A.~Sablayrolles, and H.~J{\'e}gou, ``Training data-efficient image transformers \& distillation through attention,'' in \emph{International conference on machine learning}.\hskip 1em plus 0.5em minus 0.4em\relax PMLR, 2021, pp. 10\,347--10\,357.

\bibitem{yu2022metaformer}
W.~Yu, M.~Luo, P.~Zhou, C.~Si, Y.~Zhou, X.~Wang, J.~Feng, and S.~Yan, ``Metaformer is actually what you need for vision,'' in \emph{Proceedings of the IEEE/CVF conference on computer vision and pattern recognition}, 2022, pp. 10\,819--10\,829.

\bibitem{fpn}
T.-Y. Lin, P.~Doll{\'a}r, R.~Girshick, K.~He, B.~Hariharan, and S.~Belongie, ``Feature pyramid networks for object detection,'' in \emph{Proceedings of the IEEE conference on computer vision and pattern recognition}, 2017, pp. 2117--2125.

\bibitem{FastViT}
P.~K.~A. Vasu, J.~Gabriel, J.~Zhu, O.~Tuzel, and A.~Ranjan, ``Fastvit: A fast hybrid vision transformer using structural reparameterization,'' in \emph{Proceedings of the IEEE/CVF International Conference on Computer Vision}, 2023.

\end{thebibliography}

\end{document}